\documentclass[3p, twocolumn]{elsarticle}

\usepackage{amsfonts,amssymb,amsmath,amsthm}
\usepackage{url}
\usepackage{epsfig}
\usepackage{epstopdf}
\usepackage{multirow}

\usepackage{graphicx}
\usepackage{caption}
\usepackage{subcaption}
\usepackage{lscape}
\usepackage{array}
\usepackage{threeparttable}
\usepackage{afterpage}
\usepackage{pdflscape}
\usepackage{booktabs}
\usepackage{ textcomp }
\usepackage{natbib}
\newcolumntype{C}[1]{>{\centering\let\newline\\\arraybackslash\hspace{0pt}}m{#1}}
\newcolumntype{L}[1]{>{\raggedright\let\newline\\\arraybackslash\hspace{0pt}}m{#1}}
\usepackage[switch]{lineno}
\usepackage{etoolbox}
\appto\TPTnoteSettings{\footnotesize}

\journal{Artificial Intelligence in Medicine}

\begin{document}

\begin{frontmatter}



%\title{An Affordable Telecardiology for Early Warning Generation: A Dictionary Based Approach to Detect Premature Ventricular Contractions}

\title{Dictionary-based Monitoring of Premature Ventricular Contractions: \\
An Ultra-Low-Cost Point-of-Care Service}

\author[iitee]{Bollepalli~S.~Chandra\corref{cor1}}
\ead{bschandra@iith.ac.in}
\author[iitma]{Challa~S.~Sastry}
\ead{csastry@iith.ac.in}
\author[mc]{Laxminarayana Anumandla}
\ead{laxmin56@gmail.com}
\author[iitee]{Soumya~Jana}
\ead{jana@iith.ac.in}
\address[iitee]{Dept. of Electrical Engineering, Indian Institute of Technology Hyderabad, India - 502285.}
\address[iitma]{Dept. of Mathematics, Indian Institute of Technology Hyderabad, India - 502285.}
\address[mc]{Dept. of Cardiology, Maxcare Hospital, Warangal, India - 506001}
\cortext[cor1]{Corresponding author}	
%
%\author[EE]{Bollepalli~S.~Chandra}
%\ead{bschandra@iith.ac.in}
%\author[MA]{Challa~S.~Sastry}
%\ead{csastry@iith.ac.in}
%\author[EE]{S.~Jana}
%\ead{jana@iith.ac.in}
%
%\address[EE]{Dept. of Electrical Engineering, IIT Hyderabad, India-502205}
%\address[MA]{Dept. of Mathematics, IIT Hyderabad, India-502205}

\begin{abstract}
While cardiovascular diseases (CVDs) are prevalent across economic strata, the economically disadvantaged population is disproportionately affected due to the high cost of traditional CVD management, involving consultations, testing and monitoring at medical facilities. Accordingly, developing an ultra-low-cost alternative, affordable even to groups at the bottom of the economic pyramid, has emerged as a societal imperative. Against this backdrop, we propose an inexpensive yet accurate home-based electrocardiogram (ECG) monitoring service. Specifically, we seek to provide point-of-care monitoring of premature ventricular contractions (PVCs), high frequency of which could indicate the onset of potentially fatal arrhythmia. Note that a traditional telecardiology system acquires the ECG, transmits it to a professional diagnostic center without processing, and nearly achieves the diagnostic accuracy of a bedside setup, albeit at high bandwidth cost.  In this context, we aim at reducing cost without significantly sacrificing reliability. To this end, we develop a dictionary-based algorithm that detects with high sensitivity the anomalous beats only which are then transmitted. We further compress those transmitted beats using class-specific dictionaries subject to suitable reconstruction/diagnostic fidelity. Such a scheme would not only reduce the overall bandwidth requirement, but also localizing anomalous beats, thereby reducing physicians' burden. Finally, using Monte Carlo cross validation on MIT/BIH arrhythmia database, we evaluate the performance of the proposed system. In particular, with a sensitivity target of at most one undetected PVC in one hundred beats, and a percentage root mean squared difference less than 9\% (a clinically acceptable level of fidelity), we achieved about 99.15\% reduction in bandwidth cost, equivalent to 118-fold savings over traditional telecardiology. In the process, our algorithm outperforms known algorithms under various measures in the telecardiological context.

\end{abstract}


\end{frontmatter}

%\newpage
%\linenumbers	
%%n

\section{Introduction} \label{sect1}

Cardiovascular diseases (CVDs) are a leading cause of death across economic strata \cite{WHO}. Hence a crucial healthcare objective consists in managing those diseases. In this regard, electrocardiogram (ECG) signals acquired from subjects often play a vital role. Specifically, continuous ECG monitoring is central to early diagnosis and improved clinical outcome in certain scenarios. However, such monitoring at a professional facility is often unaffordable to economically disadvantaged individuals due to high cost, low availability and other barriers. Against this backdrop, home-based point-of-care (POC) monitoring assumes significance. In this paper, we propose a POC monitoring service that is highly affordable.

Symptoms indicating CVDs often manifest sporadically. Consequently, to detect deviations from the normal sinus rhythm, subjects should ideally be monitored continuously. Especially, for patients who have suffered myocardial infarction (MI), or developed left ventricular dysfunction (LVD), continuous monitoring has proven essential in promptly detecting sudden deterioration in cardiac functions, and hence preventing mortality \cite{podrid}. The aforementioned as well as various related conditions are associated with premature ventricular contractions (PVCs) that briefly interrupt the normal rhythm of the heart \cite{gertsch}. Although PVCs occur in healthy individuals as well, high frequency of PVCs is known to foretell serious arrhythmic conditions \cite{hinkle}, and significantly correlate with events of mortality \cite{ng}. In short, accurate detection of PVCs assumes clinical significance in stratifying high risk patients, and predicting medical emergencies. In this context, we propose a novel personalized service to monitor the PVC burden.

In particular, we seek to develop a POC service that would appeal to the economically disadvantaged. Worldwide, about 1.2 billion individuals live on less than US\$ 1.25 per day and have little discretionary income \cite{poverty}. To such individuals, the cost of professional monitoring could often be prohibitive. Further barriers to quality care could include travel and hospital expenses. Fortunately, high penetration of mobile phones even in remote communities has mitigated such barriers in certain scenarios \cite{west}. In the present case, can the mobile network be leveraged to provide reliable PVC monitoring at an attractive cost to the communities living at the bottom of the economic pyramid \cite{BOP}?

\begin{figure}[t!]
	\centering
\begin{tabular}{c}
		\epsfig{file=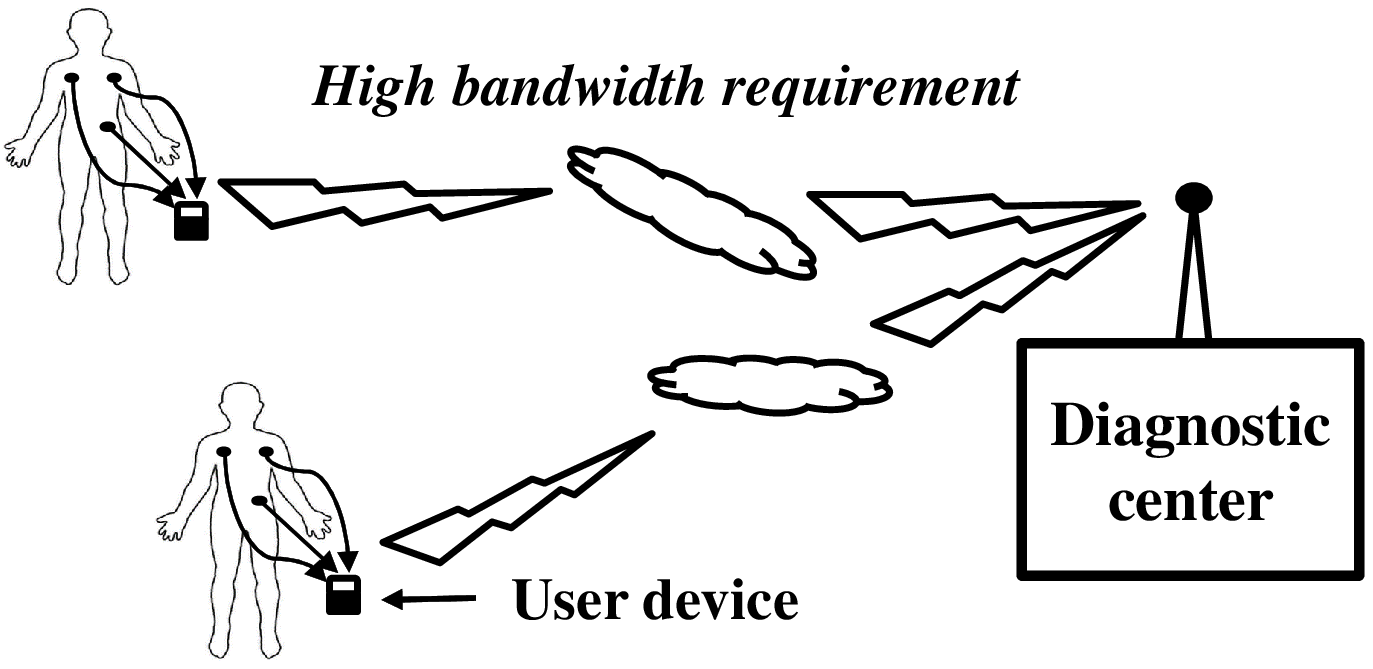, width=3in} \\
(a)		\\
\\
\epsfig{file=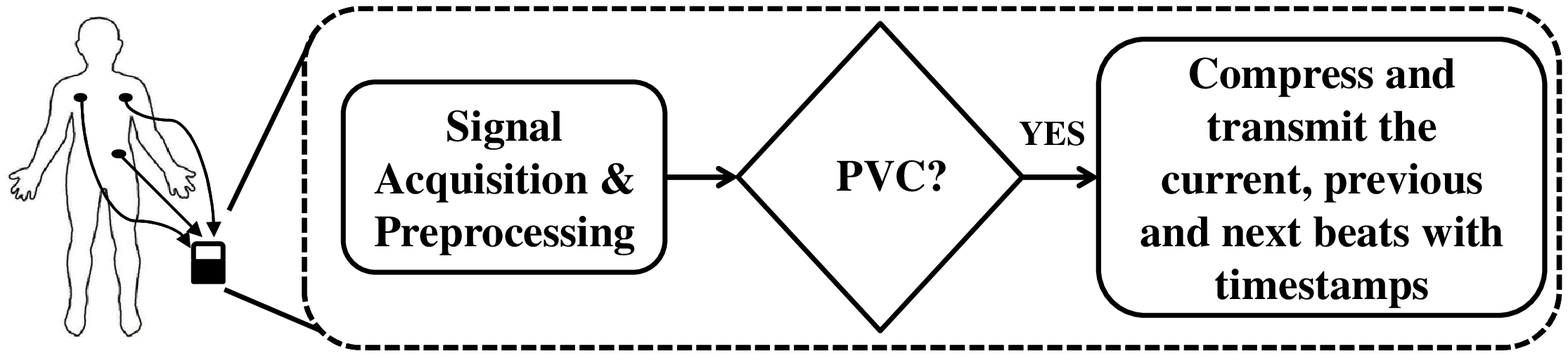, width=3in} \\
(b)		\\

\end{tabular}
\caption{(a) Traditional telecardiology architecture; (b) Schematic of proposed PVC detection and compressed beat-trio communication performed in the user device.}
\label{arch}
\end{figure}

In response, we take a frugal engineering approach \cite{frugal}, and propose an ultra-low-cost POC service. As depicted in Figure \ref{arch}a, a conventional telecardiology system simply records user ECG and transmits it to a diagnostic center staffed by medical professionals, where anomalies are manually detected and medical intervention is initiated, when necessary. Traditionally, ECG signals are transmitted unaltered, resulting in perfect accuracy (subject only to human error), albeit with the attendant high bandwidth cost and without localizing potentially anomalous beats. To reduce cost, we propose a new telecardiology paradigm, depicted in Figure~\ref{arch}b, where each user is equipped with a heartbeat classifier that detects anomalous beats, and then compresses and transmits only those anomalous beats and delimiting neighbors (forming beat-trios) along with timestamps. Such a system not only reduces the bandwidth requirement but also presents to medical professionals only those beats that warrant closer inspection, thereby potentially improving the responsiveness of the diagnostic center. 

In this framework, system design involves a tradeoff among three quantities: (i) classifier sensitivity (the fraction of PVC beats correctly identified), which we take as the reliability criterion, (ii) the fidelity of reconstructed signal at the diagnostic center, which determines the ability of experts to authenticate algorithmic classification, and (iii) the transmission bandwidth, which dictates the operating cost. To ensure accurate clinical outcome, one desires both high reliability (classifier sensitivity) and high fidelity. At the same time, one also seeks low transmission bandwidth in order to operate at a low cost. The main difficulty arises due to the complex three-way tradeoff among the above quantities. In particular, the bandwidth usage increases with sensitivity and decreases with specificity, while sensitivity and specificity themselves exhibit a nonlinear inverse inter-relationship dependent on reconstruction fidelity. The above quantities are further affected by signal compression. In this paper, we study the said tradeoff, and propose a natural design framework for telecardiology systems.

None of the individual tasks, namely, classification and compression, is new in the field of ECG signal processing. In fact, numerous algorithms have been reported specifically for PVC detection. Examples include machine learning algorithms, such as mixture of experts \cite{palreddy}, linear and quadratic discriminant analyses \cite{chazal, Llamedo}, support vector machine \cite{melgani}, and artificial neural networks  \cite{jiang, ince, zzhang, kiranyaz, bortolan}. However, such algorithms have not been designed to achieve compression as well, and are not optimized in the high-sensitivity regime. On the other hand, reported ECG compression algorithms are based on techniques, ranging from the classical time and transform domain methods \cite{tai, zigel1}, to the recent overcomplete dictionary learning \cite{fira1, adamo}. Yet, those algorithms too were not designed to achieve the desired high-sensitivity classification. Against this backdrop, we propose a dictionary-based method that attempts to retain the best of both worlds, and achieve the desired classification and the compression goals simultaneously. The proposed approach, however, is different from the (symmetric) joint classification/reconstruction framework \cite{jana}, where a combination of classification and (class-oblivious) reconstruction indices is minimized subject to a rate constraint. Instead, our setup is inherently class-asymmetric as signals detected as normal (excepting delimiting beats) are discarded, i.e., compressed to zero bits, while signals detected as PVCs are compressed at a certain rate so as to meet a target diagnostic fidelity criterion.

To this end, we propose to train separate overcomplete dictionaries for the respective classes of normal and PVC beats using labeled data. Specifically, each test beat is approximated as a linear combination of the columns of each dictionary. Intuitively, a signal should admit sparse representation only in the dictionary of the matching class. In accordance, the ratio of sparsity of representation in each dictionary is computed, and a suitable class is assigned by comparing that ratio to a threshold. The sensitivity level is then tuned to a desired level by varying such threshold. Here, the sparsity of representation is dictated by the desired fidelity of reconstruction, and in turn determines the degree of compression. Although low reconstruction fidelity would result in high sparsity (hence high compression), the resulting representation would also tend to miss the information necessary for accurate classification. Interestingly, high-fidelity regime may not guarantee accurate classification either. As the signal approximation becomes increasingly accurate, the representations based on rival dictionaries would decrease in sparsity, which in turn leads to poor classification. Consequently, our task involves choosing suitable level of fidelity so that high sensitivity and high compression are both achieved.

The efficacy of the proposed scheme is demonstrated on the standard MIT/BIH arrhythmia database using Monte Carlo cross validation (MCCV). Presently, we confine to PVC beat detection, however, the same framework can be extended to detection of other as well as multiple classes of anomalies. Specifically, at a high-sensitivity target of 99\% (i.e., no more than one undetected PVC beat in one hundred), using only classification and only compression, we respectively reduced the bandwidth requirement to 42.4\% and 2.0\% compared to the original. Using both classification and compression, we required a bandwidth of only 0.85\% of the original, which translates to 118-fold savings in the operating cost, and an ultra-low-cost solution. Finally, we compared results obtained by our technique with those obtained using existing algorithms, and demonstrated the criticality of the proposed high-sensitivity approach in realizing practical ultra-low-cost telecardiology.
%In the course, we presented the tradeoff between bandwidth savings and reconstruction accuracy at various sensitivity targets.

Our key contributions are as follows. We
\begin{enumerate}
\item developed a dictionary-based algorithm that achieves high-sensitivity classification and high-fidelity compression;
\item demonstrated an affordable POC service based on such algorithm, and evaluated its efficacy using MCCV on the standard MIT/BIH arrhythmia database;
\item achieved 118-fold cost reduction over classical telecardiology, which improves upon the cost reduction due to known algorithms.
\end{enumerate}
The rest of the paper is organized as the following. Sec. \ref{sec:motivation} details our motivation, and identifies the key medical and social goals. In Sec. \ref{sec:sigpro}, the associated signal processing problems are formalized with necessary mathematical treatment. Dictionary-based solutions are developed in Sec. \ref{sec:soln}. Performance evaluation strategy, experimental setup and simulation results are presented in Secs. \ref{sec:perf}, \ref{sec:expt} and \ref{sec:results}, respectively. Finally, Sec. \ref{sec:disc} concludes the paper with a discussion.

\section{Motivation and Envisaged System}

\label{sec:motivation}

We begin by placing the present problem in medical and social contexts.
\subsection{Clinical Imperative}

Cardiac anomalies could be caused by various conditions that overwork and/or damage heart muscles. Continuous monitoring has often proven effective in timely detection of such anomalies. In particular, monitoring PVCs, which are an early depolarization of the myocardium originating in the ventricle \cite{gertsch}, assumes significance, even though such beats are found in subjects with as well as without structural heart diseases \cite{hinkle}. In healthy individuals, a PVC prevalence of less than $1\%$ is common, which carries no prognostic significance. In contrast, more frequent PVCs might indicate (or, lead to) structural heart diseases. Specifically, 90\% of patients experience PVCs after acute MI \cite{vlodaver}, and the risk of sudden death in such patients is related to the complexity and frequency of the PVCs. Recent studies also indicate the role of PVCs in inducing cardiomyopathy \cite{cha}. 

More generally, continuous monitoring of PVCs has proven effective in stratifying clinical risk. However, there is no clear demarcation between high and low frequencies of PVCs. Recommended lower threshold for the high-risk subjects, such as those with a history of MI or LVD varies between 10,000 and 20,000 in a 24-hour window \cite{niwano}. Another recommendation sets 10\% as the threshold PVC burden \cite{baman}. Besides frequency of PVCs, run of two or more PVCs and their complexity could also indicate an adverse heart condition \cite{ng}. Accordingly, in the present work, we propose a PVC monitoring system that detects PVC beats with high sensitivity and communicates those with high fidelity to the diagnostic center, when suitable high-risk criteria are met.

%Various studies have set 10000/20000 PVCs over 24 hours as high-frequency group \cite{niwano, kanei}. At the same time, rather than using an absolute number, a 10\% PVC burden of all the beats is set as a target to indicate critical condition \cite{baman}. Accordingly, in the present work we use the later target as a markers for serious heart conditions, and hence a basis for initiating medical intervention. 

%\begin{figure}[t]
%	\centering
%		\includegraphics{}
%{file=VB, width=3in}
%		\caption{Illustration of ECG record containing normal and ventricular beats \cite{physionet}. Beats annotated ``N" indicate normal, and ``V" indicate PVCs.}
%		\label{ecgbeats}	
%\end{figure}

%\begin{figure*}[t!]
%  \centering
%    \includegraphics[width=\textwidth]{Signal1} \\
%(a) \\
%    \includegraphics[width=\textwidth]{Signal_processed1} \\
%(b)
%\caption{A one minute ECG signal (a) In conventional setting (b) proposed beat-trio setting}
%\label{signal}
%\end{figure*}

%\begin{figure}[t]
%	\centering
%		\includegraphics{}
%{file=VB, width=3in}
%		\caption{Illustration of ECG record containing normal and ventricular beats \cite{physionet}. Beats annotated ``N" indicate normal, and ``V" indicate PVCs.}
%		\label{ecgbeats}	
%\end{figure}

\begin{figure*}[t!]
  \centering
    \includegraphics[width=\textwidth]{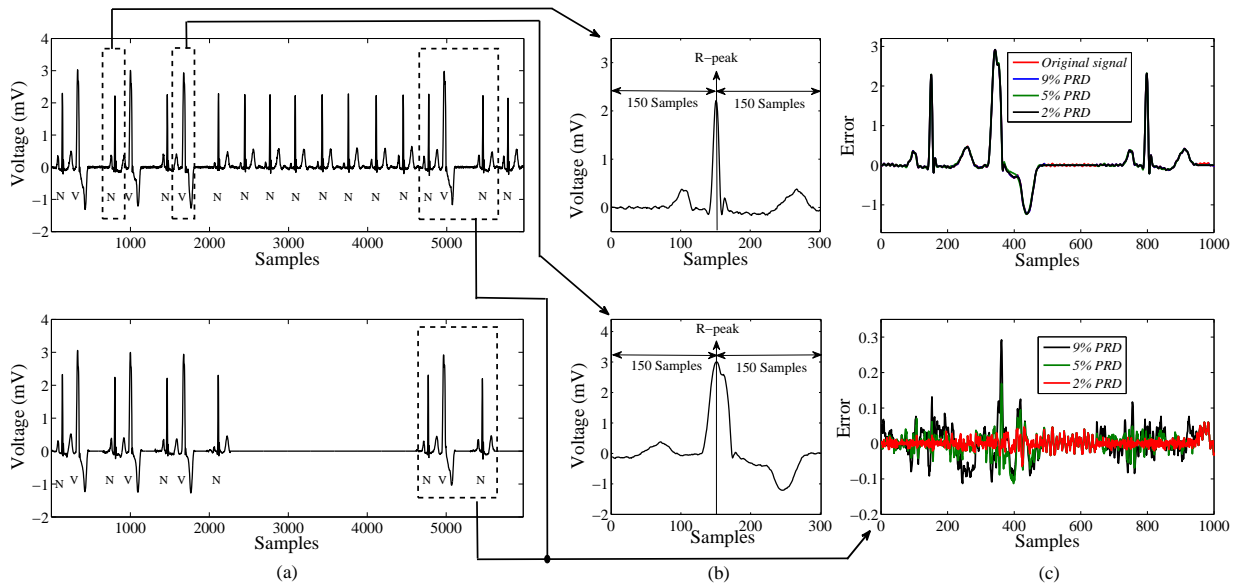} \\
\caption{(a) An entire ECG snippet (top), and only beat-trios (bottom); (b) Beat vectors for a normal beat (top), and a PVC beat (bottom); (c) Reconstructed beat-trio at various $PRD$ levels (top), and corresponding reconstruction errors (bottom).}
\label{signal}
\end{figure*}

\subsection{Technological Imperative}

A conventional telecardiology system, depicted in Figure \ref{arch}a, acquires and transmits entire user ECG to the diagnostic center. Such a system not only utilizes the available bandwidth in an inefficient manner but also burdens the medical professional with processing the entire record to identify anomalies. In this framework, telephone based ECG transmission and associated clinical experience were investigated decades ago \cite{scheidt}. With growing ubiquity of mobile phones in recent years, cellular network based as well as ZigBee based wireless systems have been developed \cite{varshney, huang,leehj}. Yet, despite technological progress, the inefficient telecardiology architecture has largely avoided scrutiny. In this backdrop, we propose a novel architecture that makes judicious use of bandwidth while assisting medical professionals by localizing the potential anomalies, without compromising on the quality of care. 

In this context, note that efforts have already been made to deliver telecardiology services in the remote and rural communities with rickety networks. In particular, a method to encode ECG signals into ASCII characters to enable communication via SMS (short message service) has been reported \cite{Mukhopadhyay}. In contrast, we assume a reliable network, which is expected to reflect the ground reality better and better with the passage of time in view of the phenomenal advancement in communication technology \cite{ICT}.

%A design framework for such systems has been presented, albeit in broader contexts \cite{baig, telecardiology}. With growing ubiquity of mobile networks and their ability to provide pervasive health care services \cite{varshney}, various mobile based telecardiology systems have been reported \cite{fang, huang}. Further, a zig-bee and mobile based remote monitoring of elderly patients has also been presented \cite{leehj}. A method to encode ECG signals to ASCII characters to communicate via SMS service has been reported in order to leverage an unreliable network  \cite{Mukhopadhyay}. In this paper, we presume a reliable network, which approximates the ground reality better and better in view of the phenomenal advancement in communication technologies \cite{ICT}.

\subsection{Social Imperative: Representative Scenario}

As alluded earlier, we seek to provide a low-cost telecardiology solution for individuals with average daily income of about US\$ 1.25. Consider an individual living at the economic threshold of this target population segment, who suffered myocardial infarction in the recent past, and was successfully treated (see \cite{steg} for various treatment options). Post treatment, monitoring PVCs over long intervals has now become a clinical priority as mentioned earlier. In this context, we shall investigate the cost associated with such PVC monitoring.

\subsubsection{Cost incurred in traditional telecardiology}
\label{traditionalCost}
Let us first estimate the cost incurred in traditional telecardiology. Here, we assume that diagnostic services are rendered free of cost. Such an assumption is realistic in various developing and underdeveloped countries, where free healthcare is dispensed from government-run facilities \cite{purohit}. This welfare paradigm is currently being extended even to the broader context of telemedicine \cite{jaroslawski}. So the cost incurred would only constitute the data transmission cost. Considering a sampling rate of 360Hz and word length of 11 bits (used in MIT/BIH arrhythmia database \cite{physionet}), one would generate about 1.78MB of data per hour. As we plan to use existing mobile networks, communicating entire data to the diagnostic center would cost about US\textcentoldstyle~27 per hour at the rate of US\textcentoldstyle~1.5 per 100KB of data usage\footnote{We use the Indian mobile data tariff of Indian rupee (INR) 1 per 100KB as representative, and an exchange rate of US\$ 1 = INR 66.7.}. At this rate, the cost of ten-hour monitoring of single channel ECG would amount to US\$ 2.7.

\subsubsection{Affordability as necessity}

In general, healthcare expenses exceeding 10\% of household spending is considered catastrophic \cite{affordability}. In the aforementioned scenario, assuming a household size of four excluding the subject (which approximates the average family size in India \cite{household}), the household income amounts to about US\$ 5 a day. Assuming zero savings, the 10-hour PVC monitoring cost of US\$ 2.7, calculated in Sec. \ref{traditionalCost}, amounts to 54\% of the household spending, and would clearly be unaffordable. In this situation, as a catastrophic health condition is expected to be detected rarely, the household could be tempted to view the monitoring expenditure as non-essential. In reality, however, timely detection of a life-threatening condition saves life with high probability, and hence periodic monitoring remains crucial for long-term survival. Hence, it becomes imperative that the monitoring cost be drastically reduced to such an affordable level that even an economically disadvantaged person would find little incentive to forego it.

\subsection{Outline of Envisaged System}

To meet the aforementioned imperative, we seek to reduce the volume of data communicated to the diagnostic center. As a means, it appears natural to compress the data before transmission. In fact, to make the system even more efficient, we propose to detect anomalous beats, and communicate a compressed version of only those beats. More precisely, we shall form beat-trios, each consisting of a PVC beat, and normal beats preceding and following it (See Figure \ref{signal}a for an illustrative example). A representative beat vector for the normal beat and the PVC beat are shown in Figure \ref{signal}b. If a PVC beat is not isolated, but a run of PVCs (two or more) occur, the normal beats preceding and succeeding the run are used as delimiters. Such beat-trios (and delimited PVC runs) will then be communicated to the diagnostic center along with the timing information. Although this scheme adds a worst-case overhead of two beats for each anomalous beat, it preserves the timing and morphological information of neighboring beats, which are known to facilitate professional diagnosis \cite{kiranyaz}. As mentioned earlier, additional bandwidth savings is achieved by transmitting the compressed version of those beats. The original and the reconstructed beat-trio signals along with reconstruction errors for various compression factors are presented in Figure \ref{signal}c. Specific details on the proposed classification and compression techniques are provided later. In summary, we envisage a low-cost system that makes efficient use of bandwidth by suitably classifying and compressing heart beats.

\section{Classification, Compression and Dictionary Learning}
\label{sec:sigpro}

As alluded earlier, signal processing in the present work involves classification and compression of ECG signals. In this section, we pose the associated engineering problems, and provide necessary mathematical preliminaries.

\subsection{ECG classification}

A desired classifier specifies two mutually exclusive and exhaustive subsets $\Gamma_1$ and $\Gamma_2$ of set $\Gamma$ of possible ECG beat $x$ as follows. Any beat $x\in\Gamma_1$ is declared normal, while any beat $ x\in \Gamma_2$ is declared a PVC. Presently, we wish to find $\Gamma_2$ (and hence $\Gamma_1$) such that for a given sensitivity $Se$, i.e., fraction of PVC beats correctly detected as PVC beats, the specificity $Sp$ i.e., fraction of normal beats correctly detected as normal beats is maximized \cite{NP}. 

Next we examine the bandwidth requirement of the aforementioned classifier, assuming that only beats detected as anomalous (PVC) are transmitted to the diagnostic center, which possesses adequate resources to validate and correct, if necessary, the class of each beat it receives. In other words, one fails to detect a PVC beat only if that beat is originally classified as normal and never transmitted. Thus the fraction of undetected PVCs, $1-Se$, inversely relates to the reliability of the overall system including the diagnostic center. Perfect reliability is achieved when $Se=1$. Denoting by $\rho$ the prevalence rate of PVCs, and taking the bandwidth requirement without classification as the reference, the fraction of actual PVC beats that are classified as PVCs equals $Se \times \rho$, and the fraction of normal beats that are mistakenly classified as PVCs is given by $ (1-Sp) \times (1-\rho)$. Thus the overall fraction of beats declared as PVC equals $(Se \times \rho + (1-Sp) (1-\rho))$. 

In the envisaged beat-trio system, assuming a worst-case scenario that each PVC beat is preceded and followed by normal beats, the (conservatively estimated) fraction $B_{cl}$ of bandwidth usage with only classification and no compression is given by
\begin{equation}
B_{cl} = 3(Se \times \rho + (1-Sp) (1-\rho)).
\label{SeSp}
\end{equation} 
Employing an ideal classifier ($Se=1$, $Sp=1$), one would require a bandwidth $B= 3\rho$, amounting to a substantial bandwidth savings (when $\rho <<\frac{1}{3}$), while ensuring perfect reliability. Unfortunately, such an ideal classifier is unrealizable. In practice, we seek to significantly save bandwidth, while still achieving high reliability.

\subsection{ECG compression}

In the same vein, assuming the signal set as composed of only normal and PVC beats with a compression ratio of $\beta_N~(\ge 1)$ and $\beta_V~(\ge 1)$, respectively, for normal and PVC beats, bandwidth usage is a function of prevalence and given by 
\begin{equation}
B_{co} = \rho \times \frac{1}{\beta_V} + (1-\rho) \times \frac{1}{\beta_N}.
\label{SeSp}
\end{equation}
Further bandwidth savings can be  achieved by employing a hybrid scheme, where beat-trios are formed around detected PVC beats, which are then compressed and communicated to the diagnostic center. Employing such a scheme, bandwidth usage diminishes to at most
\begin{equation}
B_{tr} = (Se \times \rho + (1-Sp) (1-\rho))(\frac{1}{\beta_V} + \frac{2}{\beta_N}).
\label{overall_cost}
\end{equation}

In general, the reconstruction fidelity varies inversely with compression ratio, and the tradeoff is beat-type specific. We shall measure reconstruction fidelity using the percentage root mean squared difference ($PRD$), widely used in the context of ECG:

\begin{equation} \label{eq:error} 
 PRD = \frac{\|x-\hat x\|_2}{\|x\|_2},
\end{equation}
where $x$ and $\hat x$ stand respectively for the original and the reconstructed signals \cite{sornmo}. Further, from a diagnostic perspective, a $PRD$ of no more than 9\% has been found to be ``good" (Table \ref{PRD}) \cite{wdd}. Accordingly, we set the above fidelity constraint in subsequent analysis. 

\begin{table}
\centering
    \begin{tabular}{cc}
    \toprule
    $PRD$ range  & Signal quality   \\ \midrule
    {\bf 0\% -- 2\% }  &{\bf Very good } \\
    {\bf 2\% -- 9\% }  & \bf{Good}  \\
    9\% -- 19\%  & Not good  \\
    19\% -- 60\%  & Bad       \\ \bottomrule
    \end{tabular}
\caption{Relation between $PRD$ and the diagnostic content of the ECG signal \cite{wdd}.}
\label{PRD}
\end{table}

\subsection{Dictionary-based Technique}
\label{sec:dict}
So far, we have envisaged a system with certain target classification accuracy and reconstruction fidelity. Now we require an enabling technology to achieve those targets. In this regard, we propose a dictionary-based solution. First we need mathematical preliminaries of compressive sampling and dictionary learning. 

\subsubsection{Compressive sampling paradigm}
\label{sec:cs}
Compressive sampling (CS) recovers a high dimensional sparse vector $\alpha \in \mathcal{R}^n$  from a few of its  measurements $x =\Phi \alpha, x  \in \mathcal{R}^m $, $m <n$, where $\Phi$ denotes the measurement matrix \cite{elad}. Formally, we seek to solve
\begin{equation}\label{eq:l_0}
\min_{\alpha} \| \alpha \|_0 \quad subject \quad to \quad \Phi \alpha = x,
\end{equation}
where $\| \cdot \|_0$ indicates the $l_0$ (counting) norm. In general, (\ref{eq:l_0}) is intractable. Fortunately, under certain technical conditions, solution to (\ref{eq:l_0}) remains unaltered if $\| \cdot \|_0$ is replaced by the $l_1$ norm $\| \cdot \|_1$. As $l_1$ solver, we shall use orthogonal matching pursuit (OMP) in view of its simplicity, empirical effectiveness (despite its being greedy) \cite{elad}, and relatively low computational complexity of O($m^2n$) \cite{rubinstein}.

\subsubsection{Dictionary learning}

The method of dictionary learning identifies a tunable selection of basis vectors providing sparse representation. Given a set of signals $\{ {x}_{i}\}_{i=1}^M$, $\ K$-SVD obtains the dictionary $D$ that provides the sparsest representation for each example in this set \cite{ksvd}. It involves a two-step procedure. In the first step, for a given dictionary $D$, we obtain matrix $\Psi$ with sparse columns by solving the following optimization problem:
\begin{equation} \label{eq:spudate}
\Psi = \operatorname{\arg\min}_{\Theta}  \sum_{l} \parallel \Theta_l \parallel_{1} \; subject ~ to \; X = D \Theta,
\end{equation}
where $\Theta_l$ is the $l$-{th} column of $\Theta$, and $ X$ is the matrix whose columns are $x_i$'s. Using the above $\Psi$, the pair ($D, \Psi)$ is then updated as
\begin{multline} \label{eq:dicupdate}
({\hat{D}},\hat{\Psi}) = \arg\min_{{D},\Psi} \|{X}-{D}\Psi\|_{F}^{2} ~ subject ~ to \\
 \|\Psi_{i}\|_{0} \leq T_{0}~ \forall i,
\end{multline}
where $\Psi_{i}$ denotes the $i$-{th} column of $\Psi$, $T_{0}$ the sparsity parameter, and $\|\cdot \|_{F}$ indicates the Frobenius norm. The $\ K$-SVD algorithm alternates between sparse coding (\ref{eq:spudate}), solved by an $l_1$ solver such as OMP (CS theory), and dictionary update (\ref{eq:dicupdate}) based on iterative soft-thresholding, till convergence. The complexity of learning an $m\times n$ dictionary based on $M$ training data (signals) is O($m^2nM$) \cite{rubinstein}. However, as such learning is generally performed offline, complexity of projecting a signal vector on a dictionary and finding dictionary coefficients is a more important consideration. Fortunately, that complexity is O($m^2n$), i.e., the same order as that of OMP. Consequently, the runtime complexity of both dictionary-based classification and compression algorithms is also  O($m^2n$).

\section{Proposed Dictionary-based Solution}
\label{sec:soln}
At this point, we  are ready to propose a dictionary-based solution to achieve the desired classification and compression targets. 

\subsection{Dictionary-based classification}
\label{sec:soln_class}

\begin{figure}
\centering
\includegraphics[width=3in]{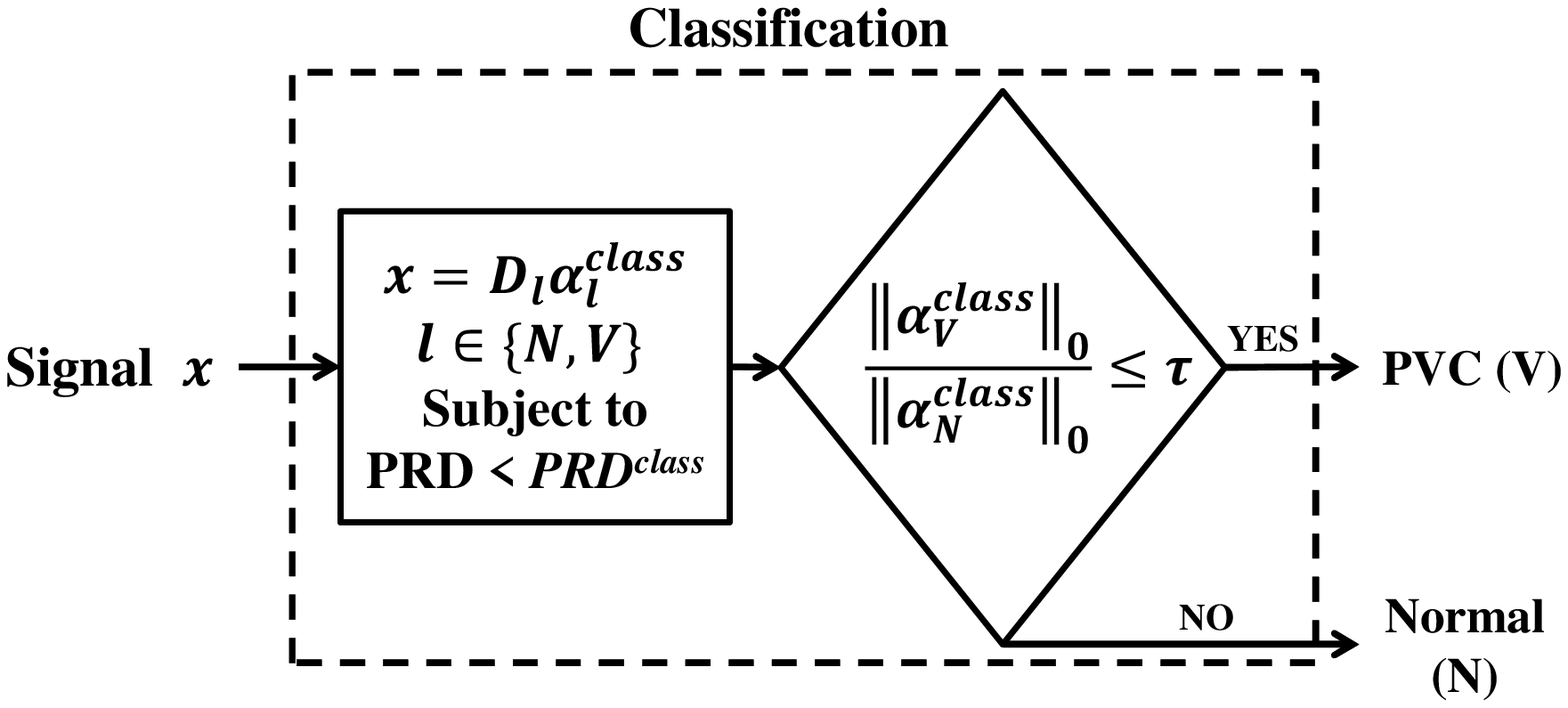}\\
(a)\\
\includegraphics[width=3in]{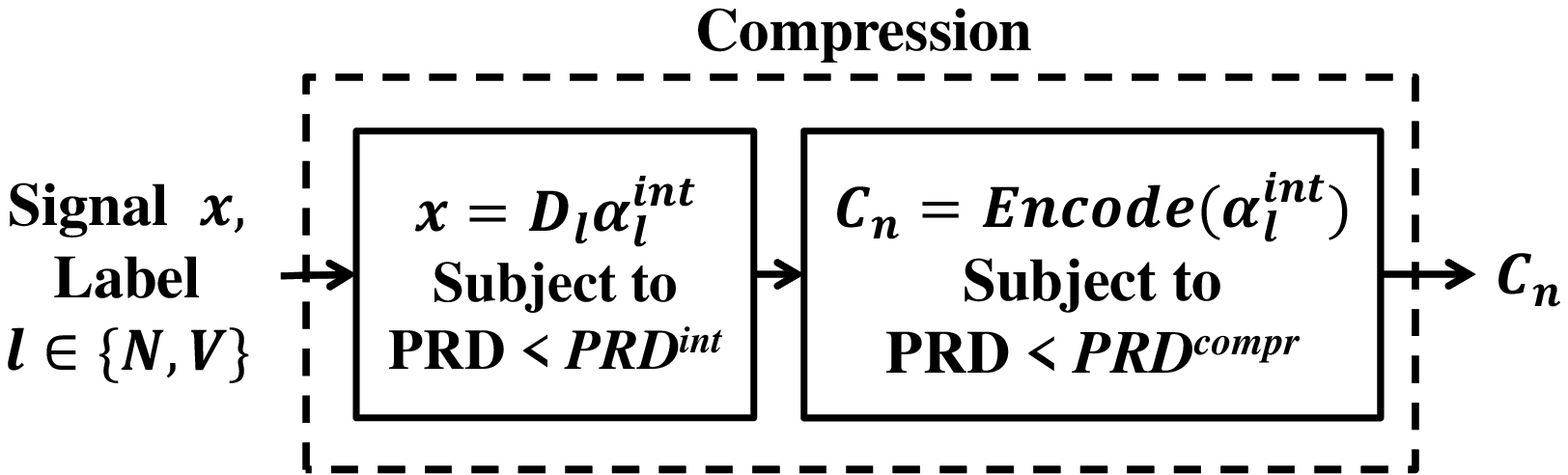}\\
(b)
\caption{Proposed dictionary-based (a) classification and (b) compression.}
\label{classifier}
\end{figure}
%\begin{figure}
%\centering
%\includegraphics[width=3.5in]{ProposedClassifier1}
%\caption{Proposed classifier.}
%\label{classifier}
%\end{figure}

Consider labeled dataset $\{ \{ {x}_{il}\}_{i=1}^{M_l} \}_{l=1}^{K}$. Here $l$ indicates the class label: $l=$``$N$'' indicates normal, and $l=$``$V$'' indicates PVC in a two class problem ($K=2$). Further, $i$ indicates beat index, taking values up to $M_l$, the number of beats present in class $l$. Based on such labeled dataset, we learn the dictionary ${D}_l \in \mathbb{R}^{m \times n}$ for class $l$. When a test beat $x$ is presented, to achieve beat classification, we first find the sparsest representation $\alpha_l$ of $x$ using each dictionary $D_l$, $l\in\{N,V\}$, by solving 
\begin{equation}\label{eq:Rl_1}
%\alpha &=& \arg\min_{\omega} \|{q}-{\hat{D}}\omega\|_{2}^{2} \; \textrm{subject to} \; \|\omega\|_{0}\leq T_{1},\\ 
\hat{\alpha}_l = \min   \| \alpha_l \|_1 \; \text{subject to} \; \|x - D_l \alpha_l\|_2 <\epsilon,
\end{equation}
where $\epsilon >0$ denotes the representation accuracy and is proportional to PRD, the normalized reconstruction fidelity.

Here, we denote by $PRD^{class}$ the target reconstruction fidelity corresponding to the classification subsystem. Operating at $PRD^{class}$, as depicted in Figure \ref{classifier}a, $x$ is marked as PVC if the ratio of $l_0$ norm (number of non-zero entries) of $\alpha_V^{class}$ to that of $\alpha_N^{class}$ is less than a suitable threshold $\tau$, and as normal otherwise. When $PRD^{class}$ is low, the signal representation tends to be non-sparse and hence our sparsity-based classification could be less accurate. Further, at high levels of $PRD^{class}$, both the dictionaries are expected to represent the signal with only a few coefficients, so that a sparser alternative is harder to pick, thereby also decreasing classification accuracy. Accordingly, we choose to operate at a suitable fidelity level that maximizes classification accuracy. Note that, for a given $PRD^{class}$, classification accuracy depends only on the ratio of sparsity of representation in rival classes, and does not require the signal to actually be reconstructed. Consequently, $PRD^{class}$ has no influence on the signal reconstruction fidelity of the overall system, and remains an internal parameter of the classification subsystem. Further, classifier performance is dictated by the choice of the threshold $\tau$. We plot receiver operating characteristic (ROC) curves for our classifier by varying $\tau$ and pick suitable operating points.

%$PRD^{int}$ corresponds to classification subsystem, 
%As the classification algorithm depends on sparsity of representation and ratio the reconstruction fidelity assumes no significance

%Note that $PRD^{int}$ corresponds only to classification subsystem remains internal to the overall system and do not influence the signal reconstruction fidelity at the diagnostic center. Further, the classifier performance is dictated by the choice of the threshold $\tau$. We plot receiver operating characteristic (ROC) curves for our classifier by varying $\tau$ and pick suitable operating point.

\subsection{Dictionary-based compression}
\label{sec:soln_compr}

%{\bf Set the goal of plotting CR versus  $PRD^{compr}$ early.}
Recall that each beat marked as anomalous, as well as each delimiting normal beat, is compressed and communicated to the diagnostic center. We intend to maximize compression ratio for a given reconstruction fidelity target $PRD^{compr}$ using a dictionary based method as shown in Figure \ref{classifier}b. Specifically, we first project the test beat on the class-specific dictionary subject to an intermediate PRD constraint $PRD^{int}$ ($\le PRD^{compr}$), and compute the corresponding dictionary coefficients, only a subset of which are expected to be non-zero. Those non-zero coefficients are subsequently quantized such that the PRD degrades enough to meet the overall constraint $PRD^{compr}$. Here, $PRD^{int}$ remains internal to compression subsystem and if we set $PRD^{int}$ to be significantly smaller than $PRD^{compr}$, the number of non-zero coefficients would be large, which would then require coarse quantization so as to increase the overall PRD sufficiently. On the other hand, if $PRD^{int}$ is set too close to $PRD^{compr}$, only a few coefficients are expected to be non-zero, which can only be quantized rather fine because of relatively small room for $PRD$ degradation. In general, $PRD^{int}$ governs the interplay between the number of non-zero coefficients and the coarseness of their quantization; however, it is not straightforward how to optimally set $PRD^{int}$ to obtain the highest compression ratio subject to $PRD^{compr}$. So, we perform a search as follows. In particular, we plot overall PRD versus compression ratio for various choices of $PRD^{int}$, and take the envelop as the plot of $PRD^{compr}$ versus compression ratio. As discussed earlier, we shall operate at $PRD^{compr}=9\%$ for each of PVC and normal classes so as to maximize signal compression while preserving the diagnostic integrity of the ECG signal.

%Next we detail our quantization scheme as well as our encoding scheme for the quantized coefficients, which in turn determines CR. We first generate a quantization table for each dictionary coefficient based on the statistics collected from training data. Specifically, we find the maximum and minimum coefficient values across all the training data, say $W_{max}$ and $W_{min}$ respectively, and quantize the range $[W_{max}, W_{min}]$ uniformly with a step size of $\Delta$. We now rank the non-zero elements of each of the coefficient vector in the descending order of the absolute value of their magnitudes. For each of the rank $i$, we find the maximum and minimum values of coefficients $W_{max}^i$ and $W_{min}^i$ respectively, across all the training vectors. We quantize the coefficients of rank $i$ to only those levels that lie in the range $[W_{max}^i, W_{min}^i]$ amounting to $\frac {W_{max}^i-W_{min}^i}{\Delta}$ quantization levels. Note that, compression performance and PRD are proportional to the step size $\Delta$.

Next we detail our quantization scheme as well as our encoding scheme for the quantized coefficients, which in turn determines compression ratio. We first generate a quantization table for each dictionary coefficient. Specifically, we rank the non-zero coefficients in the descending order of absolute magnitude. At rank $i$, we find the maximum and minimum (signed) values $W_{max}^i$ and $W_{min}^i$, respectively, and adopt uniform quantization with step size $\Delta$. Specifically, $x$ is quantized to 
\begin{equation}
Q^i (x; \Delta) = \begin{cases}
  W^i_{min} - \frac{\Delta}{2},  & x < W^i_{min}, \\
  k\Delta - \frac{\Delta}{2},  & x \in [(k-1)\Delta, k\Delta), \\
  W^i_{max} + \frac{\Delta}{2},  & x \ge W^i_{max}.
\end{cases}
\end{equation}
Note that the quantizer range depends on rank, but not the quantizer step size. Further, as the step size $\Delta$ increases, so does compression ratio as well as PRD. Finally, quantized coefficients, coefficient locations, and differential timestamps are encoded using Huffman coding algorithm based on empirical probabilities \cite{haykin}.

%Recall that the diagnostic center should be informed of both the normal ($D_N$) and PVC ($D_V$) dictionaries. Accordingly, we communicated the sparse representations $\alpha_N$ and $\alpha_V$, respectively, corresponding to the normal and the PVC beats (Figure \ref{classifier}b). We achieved compression by encoding only the quantized amplitude and the location of each of the non-zero entries of the sparse coefficient vector. Further, the time of occurrence of the beat and the beat label were also encoded. 

Finally, after quantization and encoding of dictionary coefficients, we compute the class-specific beat compression ratio $\beta$ as follows:
\begin{equation}
\label{compRatio}
\beta_l = \frac{ \text{Number of bits representing} ~ x}{\text{Number of bits representing} ~ C_l + B^{time}_l +1},
\end{equation}
where $x$ represents a beat vector, $C_l$ encodes quantized amplitude as well as location of the non-zero elements of sparse dictionary coefficients $\alpha_l$, $l\in {V, N}$, and $B^{time}_l$ represents the number of bits required to encode beat-specific timestamp. Further, one additional bit is used to encode label $l$.

\subsection{End-to-end System}
\label{sec:system}

\begin{figure}[t!]
	\centering
		\epsfig{file=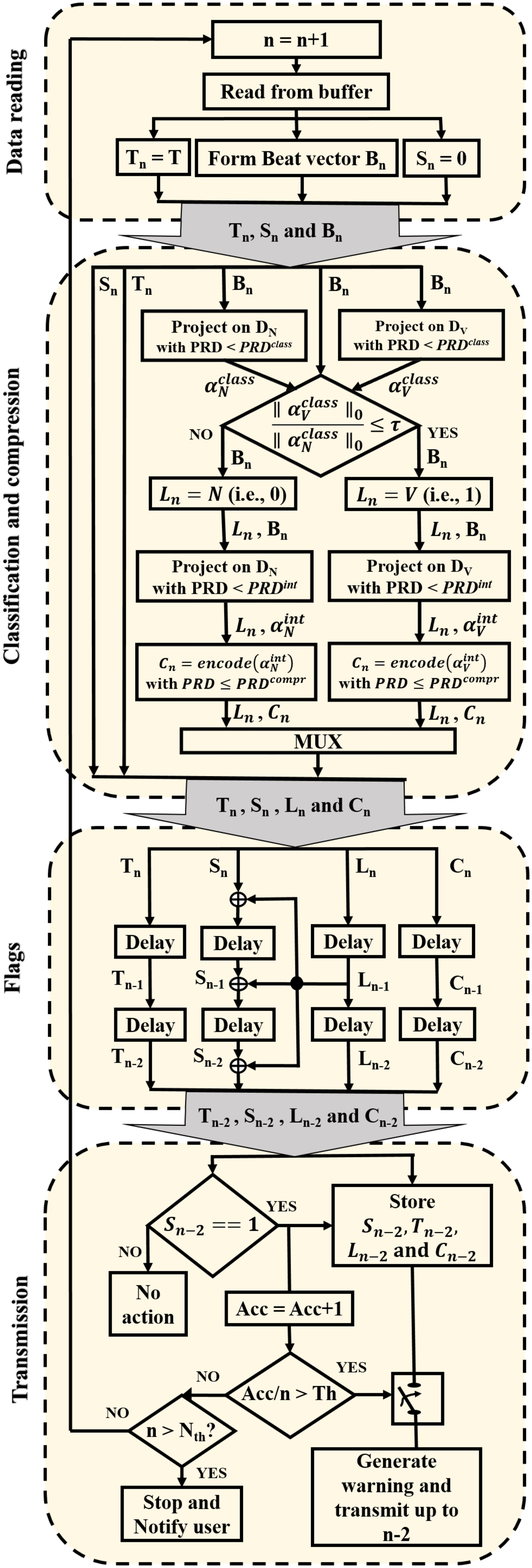, width=2.5in}
		\caption{End-to-end flowchart of the proposed service.}
		\label{fc}	
\end{figure}
%\begin{figure}[t]
%	\centering
%		\epsfig{file=FlowChart, width=3in}
%		\caption{Flowchart of the proposed method.}
%		\label{fc}	
%\end{figure}
%
%\begin{figure}[ht!]
%	\centering
%		\epsfig{file=FlowChart2, width=3in}
%		\caption{Flowchart of the proposed method.}
%		\label{fc}	
%\end{figure}

At this point, we turn to completing an end-to-end system that utilizes the classification and compression subsystems discussed so far. A flowchart of the proposed system is depicted in Figure \ref{fc}, and consists of the following modules.
%Figure \ref{fc} illustrates the flowchart of the proposed system comprising the following major modules.

\underline{\it Data reading:} We begin by acquiring ECG samples from the subject, and store those in a buffer. Simultaneously, we read stored samples from the buffer to form a beat vector $B_n$. Time $T_n$ of occurrence of corresponding beat  is recorded and the communication status flag $S_n$ is set to zero, which would later indicate whether to transmit a specific beat to the diagnostic center. 

%\underline{\it Classification/compression:} Each beat vector is projected on the pre-learned dictionaries of normal and PVC classes to obtain respective sparse representations, $\alpha_N$ and $\alpha_V$. Based on the sparsity of representation in either class, each beat is assigned a class label $L_n$ and is compressed to coefficient vector $C_n$ by encoding only the non-zero elements and corresponding indices in the class-specific representation. 

\underline{\it Classification and compression:} First, to detect anomaly, each beat vector is projected on the pre-learned dictionaries of normal and PVC classes to obtain respective sparse representations, $\alpha_N^{class}$ and $\alpha_V^{class}$, subject to $PRD \le PRD^{class}$. By comparing the ratio of sparsity of representation in either class to a threshold $\tau$, each beat is assigned a class label $L_n$. Internally, we use 0 and 1 to indicate N and V, respectively. Later, beat compression is achieved by projecting the beat vector $B_n$ on the dictionary of the chosen class, subject to $PRD \le PRD^{int}$, 
the sparse coefficient vector $\alpha_l^{int}$, $l$ = $N$ or $V$. Finally, we encode to $C_n$ only the signed magnitude of non-zero elements and corresponding indices (locations) of $\alpha_l^{int}$ subject to $PRD \le PRD^{compr}$. Here $PRD^{class}$, $\tau$, $PRD^{int}$ and $PRD^{compr}$ are design parameters.

\underline{\it Command flags:} In order to communicate only the beats detected as PVCs and delimiting normal beats, we make use of certain command flags as follows. At instance $n-1$, if a PVC beat is encountered, i.e., $L_{n-1} = V$, the communication status flags for the past, the present and the next beats are all set to 1, i.e., $S_{n-2} = S_{n-1} =  S_{n} = 1$. This is repeated after incrementing the counter n. The status flag generation logic is illustrated in Table \ref{statusflags} for three representative label sequences. As the beat label of the current beat could impact the label of the previous beat, our system would incur a delay of two beats.

\begin{table}[]
\centering
\renewcommand{\arraystretch}{1.3}
\resizebox{75mm}{!}{
\begin{tabular}{|c|ccccc|ccccc|ccccccc|}
\hline
     & \multicolumn{5}{c|}{Case-1} & \multicolumn{5}{c|}{Case-2} & \multicolumn{7}{c|}{Case-3}  \\ \hline
$L_n$ & N   & N   & ...  & N  & N  & N   & N   & V   & N   & N  & N & N & V & ... & V & N & N \\
$S_n$ & 0   & 0   &  ... & 0  & 0  & 0   & 1   & 1   & 1   & 0  & 0 & 1 & 1 &   ...  & 1 & 1 & 0 \\ \hline
\end{tabular}}
\caption{Beat labels and corresponding communication status flags for various beat sequences.}
\label{statusflags}
\end{table}

%
%
%\begin{table}[]
%\renewcommand{\arraystretch}{1.2}
%\centering
%\resizebox{75mm}{!}{
%\begin{tabular}{|c|c|c|c|}
%\hline
%     & Case-1 & Case-2 & Case-3  \\ 
%$L_n$ & N-N-...-N-N & N-N-V-N-N  & N-N-V-...-V-N-N \\
%$S_n$ & 0--0--...--0--0   & 0--0--1--0--0   &  0--0--1--...--1--0--0  \\ \hline
%\end{tabular}}
%\caption{Various beat sequences and the corresponding labels and communication status flags for each beat.}
%\label{statusflags}
%\end{table}

\underline{\it Transmission:} Finally, we communicate only the marked beats to the diagnostic center based on certain clinical considerations. Common clinical conditions requiring expert attention include: (i) Average PVC burden exceeds certain threshold over a specified interval; (ii) A run of two or more PVCs is detected. We illustrate the proposed service in Figure \ref{fc} using condition (i). Specifically, we maintain an accumulator flag $Acc$, which is incremented when an anomalous beat is detected. Once the frequency of the anomalous detections exceed the specified threshold $Th$, the user is notified and the data (with a delay of two beat intervals) are communicated to the diagnostic center. Alternatively, if the total number of beats reaches the maximum beat count $N_{th}$, monitoring stops with a notification to the user. Recall that the specific transmission logic described above is presented only as an illustration. In general, one should adopt suitable logic that embodies the desired condition.

\section{Framework for Performance Evaluation}
\label{sec:perf}

We now turn to performance evaluation of the proposed classification subsystem, the compression subsystem and the complete end-to-end system. As customary, we evaluate the classification and the compression subsystems based on the tradeoff between sensitivity (reliability) versus specificity, and compression ratio versus reconstruction fidelity, respectively. Further, we evaluate the end-to-end system in terms of bandwidth cost savings subject to clinically motivated reliability and fidelity constraints. For evaluation of various performance indices, we made use of MIT/BIH Arrhythmia database available from the PhysioBank archives \cite{physionet}. In particular, we first partition the database into training and test sets, and train a common dictionary underlying both the classification and compression subsystems. Later, we test the performance of these subsystems and the end-to-end system, and compare with the performance of reported algorithms in the telecardiological context. Clearly, the said partitioning can be carried out in large number of ways. We intend to adopt a partitioning principle that is appropriate for the underlying practical problem.

\subsection{Patient-specific Partitioning}

Traditionally, partitioning of database into training and test sets is performed either in a class-oriented or in a subject-oriented manner \cite{ye}. In the former, partitioning is based only on the heart-beat label, which allows significant amounts of data from the same patient to be represented in both training and test sets, resulting in overly optimistic performance estimates \cite{bortolan, krasteva}. In contrast, the latter seeks to account for inter-subject variability, and constitutes training and test sets with beats from distinct subsets of records, leading to an overly conservative estimate of performance \cite{zzhang}. More recently, a hybrid scheme called patient-specific training has been proposed \cite{palreddy, chazal, jiang, ince, Llamedo, kiranyaz}, in which a subject-oriented approach is taken with the following modification. A few patient-specific beats (generally, segmented from the first 5 minutes of each record) are added to the training set. Such patient-specific approach often provides a reasonable performance estimate, which is less optimistic than the performance estimated using purely class-oriented partitioning, and less conservative than that using purely subject-oriented approach.

Accordingly, we adopt a patient-specific paradigm complying with ANSI/AAMI EC57:1998 recommendation \cite{aami}, and compared with known compliant algorithms. It is worth noting that the said recommendation excludes subjects with paced beats (records 102, 104, 107 and 217 of the MIT/BIH Arrhythmia database) and partitions the remaining 44 records into training and test sets.

\begin{table*}[]
\renewcommand{\arraystretch}{1.2}
\centering
\resizebox{160mm}{!}{
\begin{tabular}{cC{6.5cm}C{6.5cm}C{2.5cm}C{2.5cm}}
\hline
\multirow{2}{*}{\begin{tabular}[c]{@{}c@{}} Partition \\ index \end{tabular}}  & \multicolumn{2}{c}{Subject ID in MIT/BIH arrhythmia database} & Number of normal beats      & Number of PVC beats    \\ \cline{2-3}  
     & training set           & test set                       & \begin{tabular}[c]{@{}c@{}}training (\%)\\ testing(\%)\end{tabular} & \begin{tabular}[c]{@{}c@{}}training (\%)\\ testing(\%)\end{tabular} \\ \hline \hline
   Partition-1      &            100, 105, 106, 108, 109, 111, 114, 116, 118, 119, 121, 123 and 124              &        200, 201, 202, 203, 205, 207, 208, 209, 210, 213, 124, 215, 219, 221, 223, 228, 230, 231, 233 and 234           &  \begin{tabular}[c]{@{}c@{}}25360 (37\%) \\ 43130 (63\%)\end{tabular} & \begin{tabular}[c]{@{}c@{}}1281 (18.3\%) \\ 5727(81.7\%)\end{tabular} \\

Partition-2                      &    100, 101, 103, 105, 106, 108, 109, 111, 112, 113, 114, 115, 116, 118, 119, 121, 122, 123 and 124             &          200, 201, 202, 203, 205, 207, 208, 209, 210, 212, 213, 214, 215, 219, 220, 221, 222, 223, 228, 230, 231, 232, 233 and 234     &  \begin{tabular}[c]{@{}c@{}}39582 (43.9\%) \\ 50499 (56.1\%)  \end{tabular} & \begin{tabular}[c]{@{}c@{}}1281 (18.3\%)  \\ 5727 (81.7\%) \end{tabular}   \\

Partition-3                      &     101, 106, 108, 109, 112, 114, 115, 116, 118, 119, 122, 124, 201, 203, 205, 207, 208, 209, 215, 220, 223 and 230             &     100, 103, 105, 111, 113, 117, 121, 123, 200, 202, 210, 212, 213, 214, 219, 221, 222, 228, 231, 232, 233 and 234  &   \begin{tabular}[c]{@{}c@{}} 45798 (50.8\%) \\ 44283 (49.2\%) \end{tabular} & \begin{tabular}[c]{@{}c@{}}3788 (54\%) \\ 3220 (46\%)\end{tabular} \\

Partition-4                      &      105, 106, 108, 109, 111, 116, 118, 124, 200, 201, 202, 203, 205, 207, 209, 210, 212, 214, 215, 223, 228 and 232          &             100, 101, 103, 112, 113, 114, 115, 117, 119, 121, 122, 123, 208, 213, 219, 220, 221, 222, 230, 231, 233 and 234  &  \begin{tabular}[c]{@{}c@{}}45575 (50.6\%) \\ 44506 (49.4\%) \end{tabular} & \begin{tabular}[c]{@{}c@{}} 4008 (57.2\%) \\ 3000 (42.8\%)   \end{tabular}  \\ \hline
                                         
\end{tabular}}
\caption{Various dataset partitions.}
\label{data_partition}
\end{table*}

%{\bf Numerical balance between training and test sets.}

\subsection{Hand-picked Partitioning} Even patient-specific partitions are numerous. Traditionally, one research group would hand-pick one such partition based on subjective criteria. Examples include Partition-1, Partition-2 and Partition-3 given in Table \ref{data_partition}. In each case, the total (and relative) numbers of normal and PVC beats are mentioned for each of the training and the test sets. In particular, Partition-1 considers only those records that contain at least one PVC beat. Here, the training and the test sets consist of 13 records with indices in the range 100--124 and 20 records with indices in the range 200--234, respectively \cite{palreddy}. Such partition has relatively a small faction (18.3\%) of PVC beats for training.
Subsequently, Partition-2 generalizes Partition-1 by considering all records with indices in the range 100--124 for training, and those with indices in the range 200--234 for testing, without paying attention to occurence of PVC beats \cite{jiang}. Though such partitioning improved numerical balance between training and test sets of normal class, PVC class still remain biased, as in the Partition-1. Finally, Partition-3 possesses the property that training and test sets enjoy approximately equal representation from the rival classes of beats \cite{chazal}. Generally, proposers of a specific algorithm tend to pick a partition that maximizes the algorithmic performance. In this vein, we observed high performance of our classification algorithm for a certain evenly split Partition-4, while adhering to the desired numerical balance (Table \ref{data_partition}). We dub as Proposal-1 the proposal to use our algorithm on Partition-4. However, comparison among algorithms in terms of peak performance observed for a hand-picked partition enjoys limited fairness and may not correlate well with user experience.

\subsection{Randomized Partitioning with Even Split} In response, we advance another Proposal-2, wherein performance is averaged over admissible evenly split partitions. In particular, 22 of 44 subjects' data are randomly chosen for training, and the remaining subjects' data for testing. Further, admissible partitions maintain the numerical balance that 45\%-55\% of the total PVC beats belong to the test set. Here, as the total number of admissible partitions is extremely large ($>10^{11}$), we adopted Monte Carlo cross validation (MCCV) approach, wherein the performance is averaged over multiple (100, in our case) random partitions. Compared to Proposal-1, the randomization in Proposal-2 more satisfactorily accounts for the unseen patient data encountered in practice.

\subsection{Randomized Partitioning with Training Set Larger than Test Set} In our home-based PVC monitoring context, one possesses voluminous historical data, and a few potential subjects to cater. Consequently, partitioning the present database such that the training set is larger than the test set appears more realistic compared to the even split seen in Proposal-1 and Proposal-2. Accordingly, we modify Proposal-2 such that each partition under consideration has 40 subjects for training and 4 for testing, and call the new Proposal-3. In Proposal-3, we also update the numerical balance between PVC and normal beats so that only those partitions, where the test set accounts for 10\% -- 20\% of the total number of PVC beats, are considered. As admissible partitions still number a large 42,294, we adopt the MCCV approach over 100 randomly chosen partitions as earlier. Of course, historical (training) data should in practice be given even more weightage over test data in view of the overwhelming preponderance of the former. However, we settle for the above split in view of the limited size of the dataset at hand.

\subsection{Recommendation}

In summary, Proposal-1 represents the peak performance, which is overly optimistic and should not be used for practical guidance. Proposal-2 provides average performance, which is more satisfactory than the peak performance in certain sense, and helps highlight the significant gap between the two. Yet, the even split in Proposal-2 does not reflect the preponderance of historical data, and hence is too conservative to guide practical design. We recommend performance figures corresponding to Proposal-3, incorporating both a more realistic split and randomization, as a (slightly conservative) design guide. 

%As customary, we report compression performance as a tradeoff between compression ratio and PRD \cite{tai, zigel1, fira1, adamo}. However, as the choice of aforementioned partition determines the class-specific dictionaries, the said tradeoff is also a function of the same partition. Accordingly, continuing in the spirit of the preceding section, we report compression performance for the aforementioned proposals P1, P2 and P3. 

%\input{DictClassifier}

%\begin{figure*}[ht]
%\centering
%\includegraphics[width=7.5in]{Sig_err}
%\caption{Original and approximated signals with corresponding error}
%\label{errSignal}
%\end{figure*}

\section{Experimental Setup}
\label{sec:expt}

At this point, we conduct simulation experiments to demonstrate the efficacy of the proposed system. First we describe the experimental setup.

\subsection{Preprocessing}

%We made use of MIT/BIH Arrhythmia database available from the PhysioBank archives for our simulations \cite{physionet}. 
Recall that the adopted MIT/BIH Arrhythmia database consists of 30-minute excerpts of two channel ambulatory ECG recordings of 48 subjects \cite{physionet}. Each channel collects 360 samples per second with a dynamic range of 10 mV peak-to-peak, and digitized to 11-bit words. Further, each beat is annotated per accepted clinical practice. For our experiments, we used the modified limb lead II (MLII) channel only.

On each record, we performed the following steps. First, the baseline wander was removed using two median filters of respective window sizes 200ms and 600ms in a sequential manner \cite{chazal}. Next the annotated R-peak location in each beat was noted, and 150 samples before, 150 samples after and the R-peak sample were collected in a vector of length 301 \cite{fira1}. Such a signal vector included most of the information contained in one heart cycle. Currently, we considered only PVC and normal beats (Figure \ref{signal}b). Although signal vectors chosen in this manner sometimes overlapped, individual beats still preserved morphological information essential for clinical diagnosis. These signal vectors were used for training dictionaries, and will be called beats from now on for the sake of simplicity.

%
%\begin{figure*}[ht]
%\begin{centering}
%
%\begin{tabular}{cccc}
%\includegraphics[width=3.8cm]{RecSig_Rsq91} & \includegraphics[width=3.8cm]{RecSig_Rsq95} & \includegraphics[width=3.8cm]{RecSig_Rsq98}  & \includegraphics[width=3.8cm]{RecSig_Rsq99}\\
%\includegraphics[width=3.8cm]{RecErr_Rsq91}  & \includegraphics[width=3.8cm]{RecErr_Rsq95} & \includegraphics[width=3.8cm]{RecErr_Rsq98} & \includegraphics[width=3.8cm]{RecErr_Rsq99}\\
%(a) & (b) & (c) & (d)\\
%\end{tabular}•
%\caption{Plot of original and recovered signal and their corresponding error for PRD of 9\%, 5\%, 2\% and 1\% respectively}
%\label{rec}
%\end{centering}
%\end{figure*}

\begin{figure}[t]
\begin{centering}
\begin{tabular}{cc}
\includegraphics[width=3.5cm]{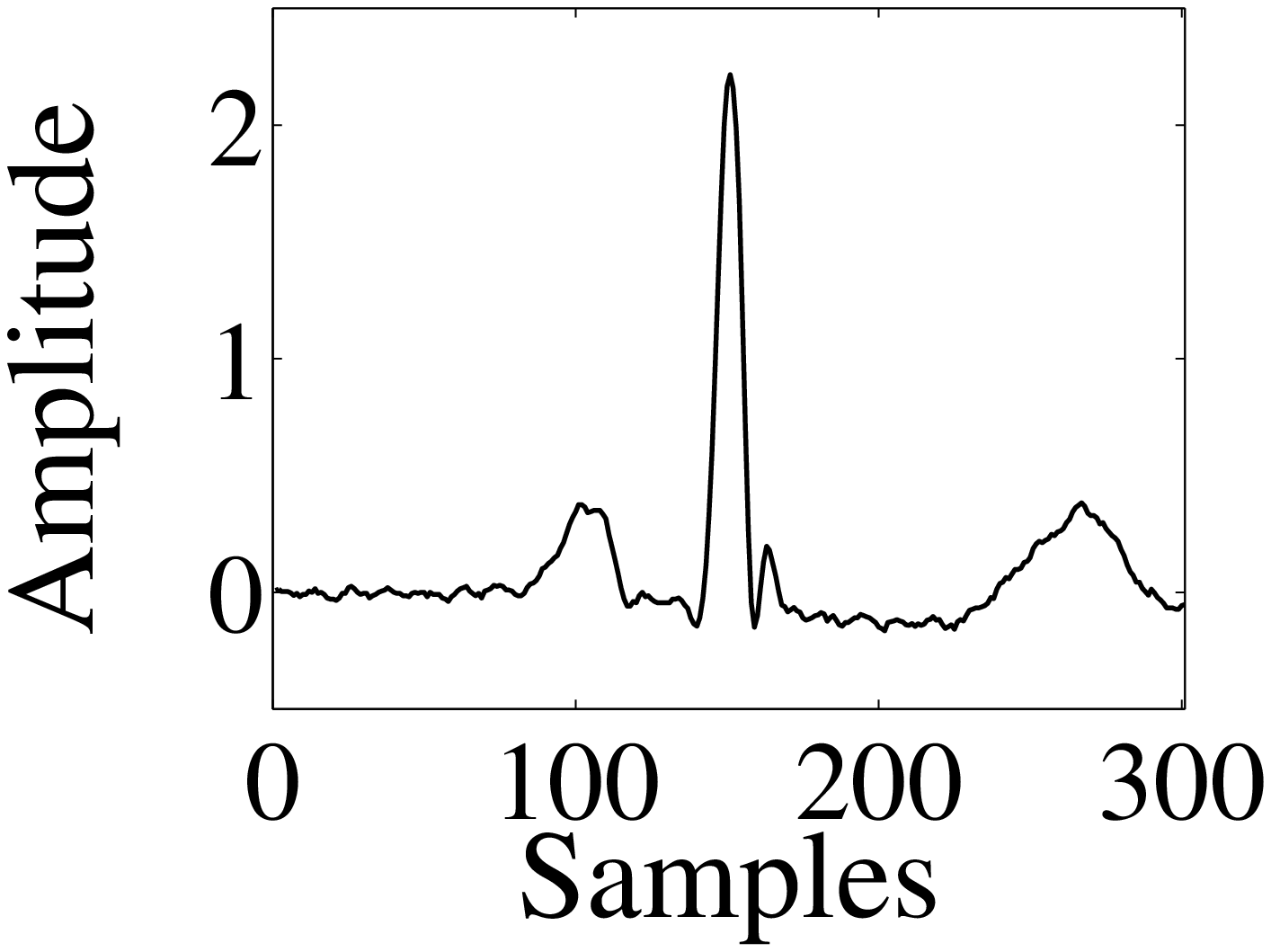} & \includegraphics[width=3.5cm]{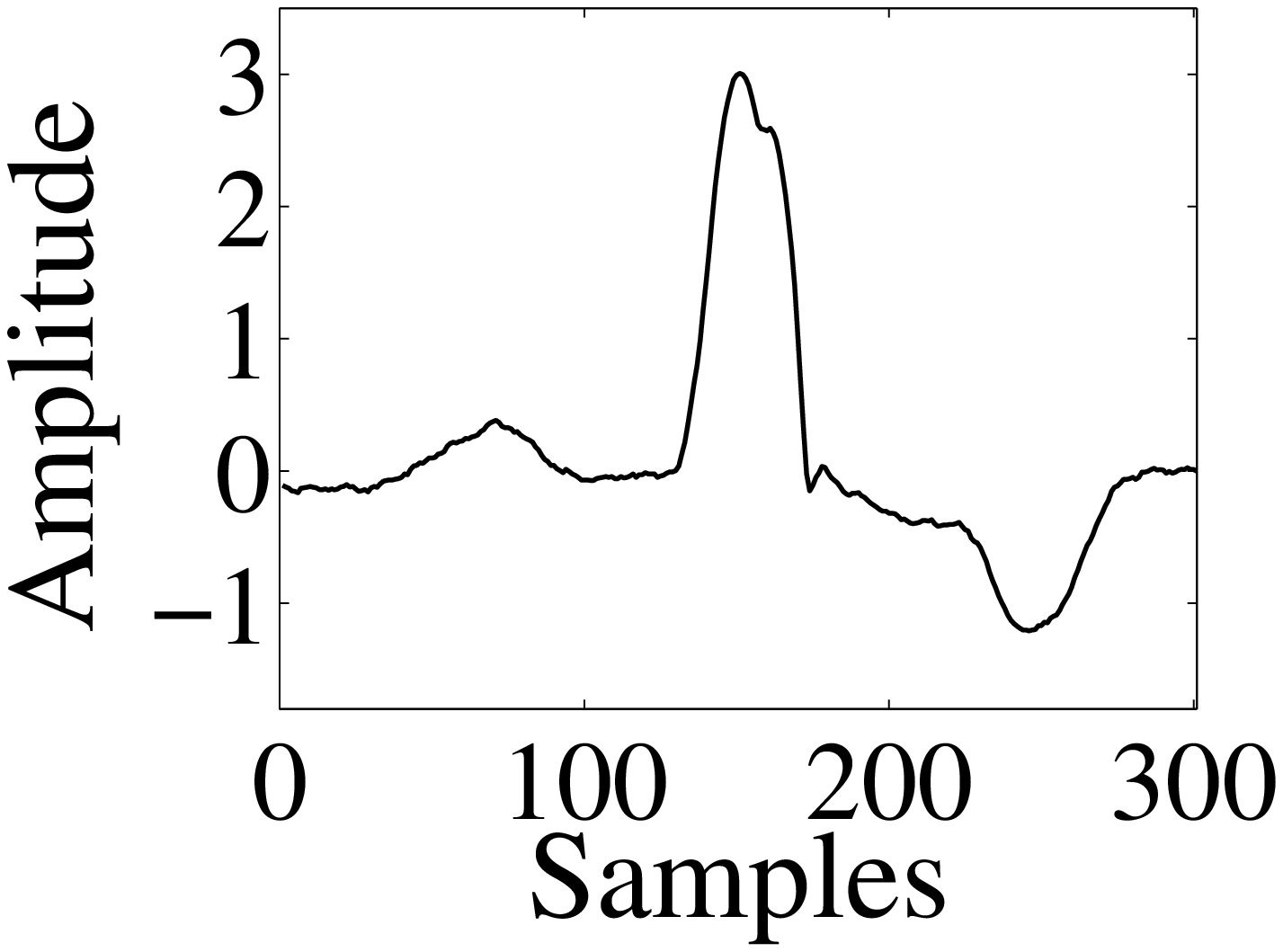}\\
(a) & (b) \\
\includegraphics[width=3.5cm]{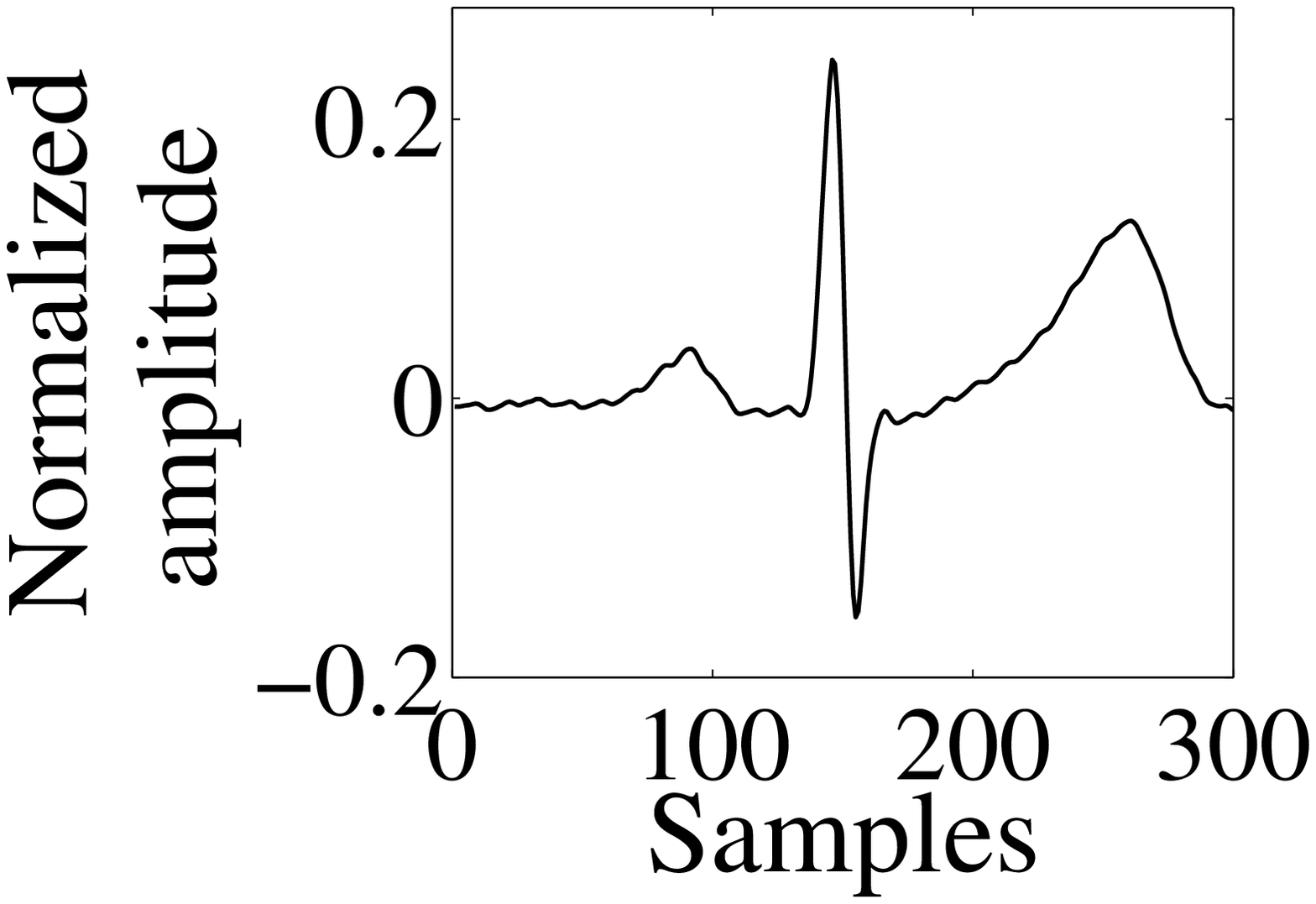} & \includegraphics[width=3.5cm]{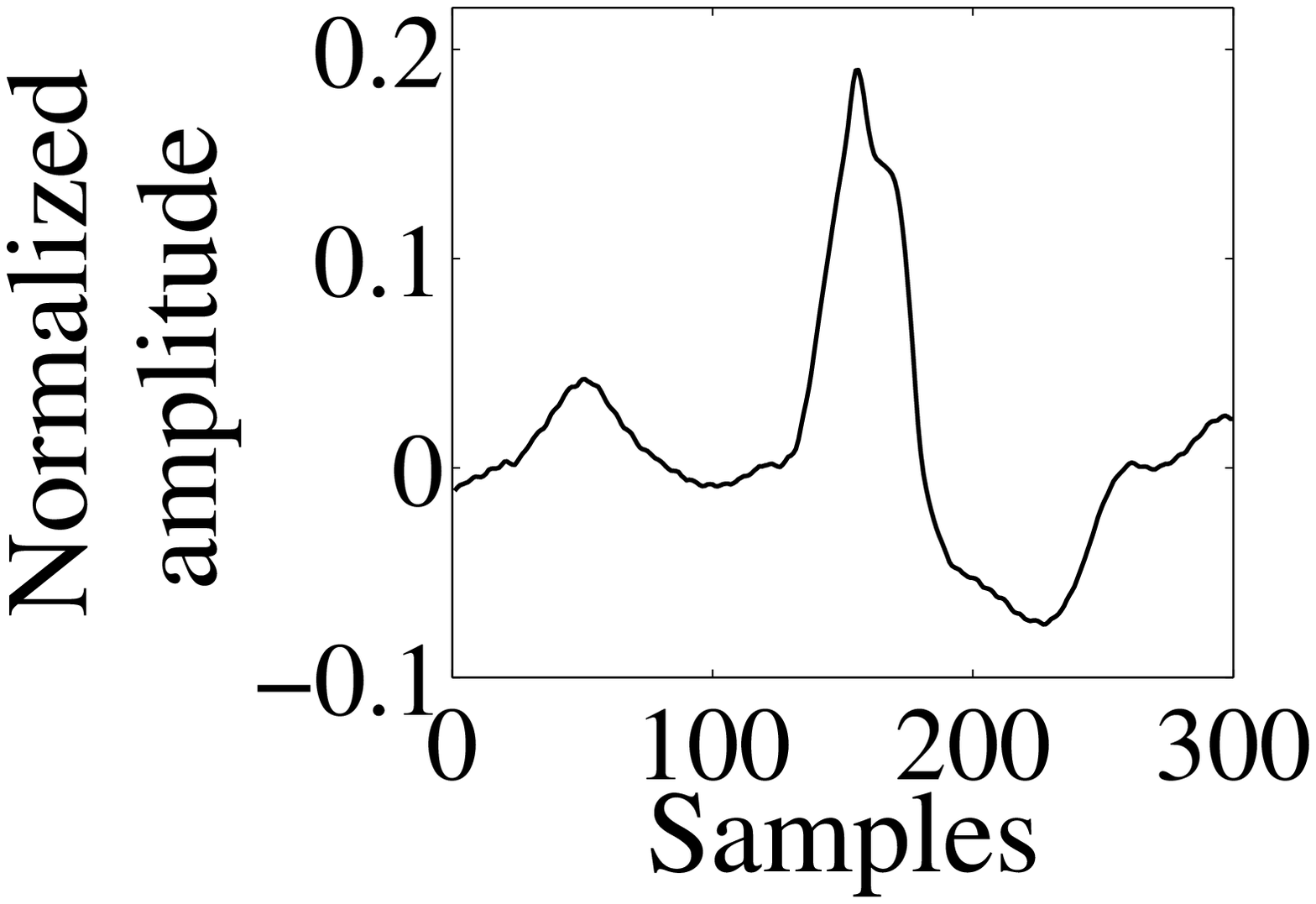}\\
\includegraphics[width=3.5cm]{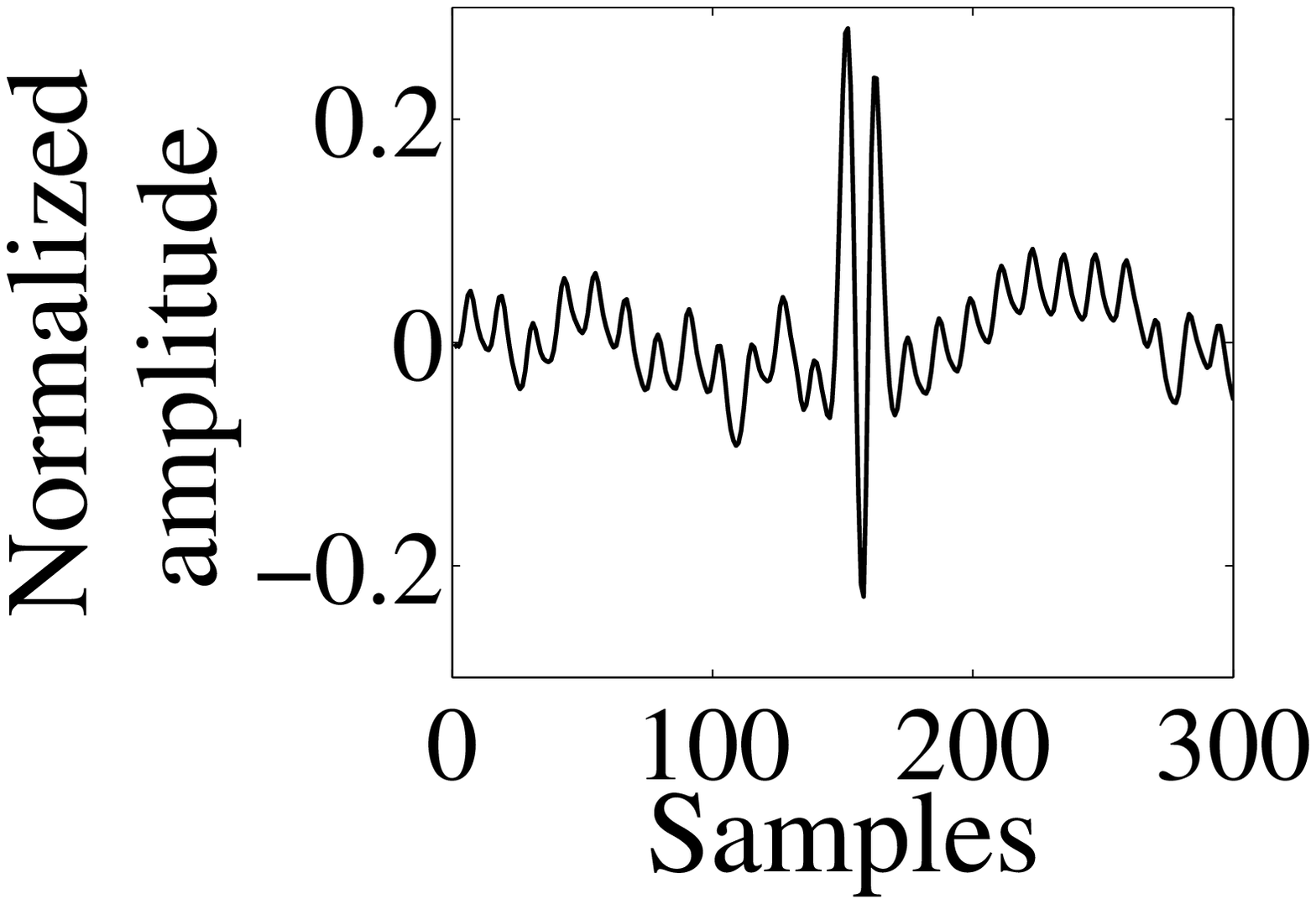}  & \includegraphics[width=3.5cm]{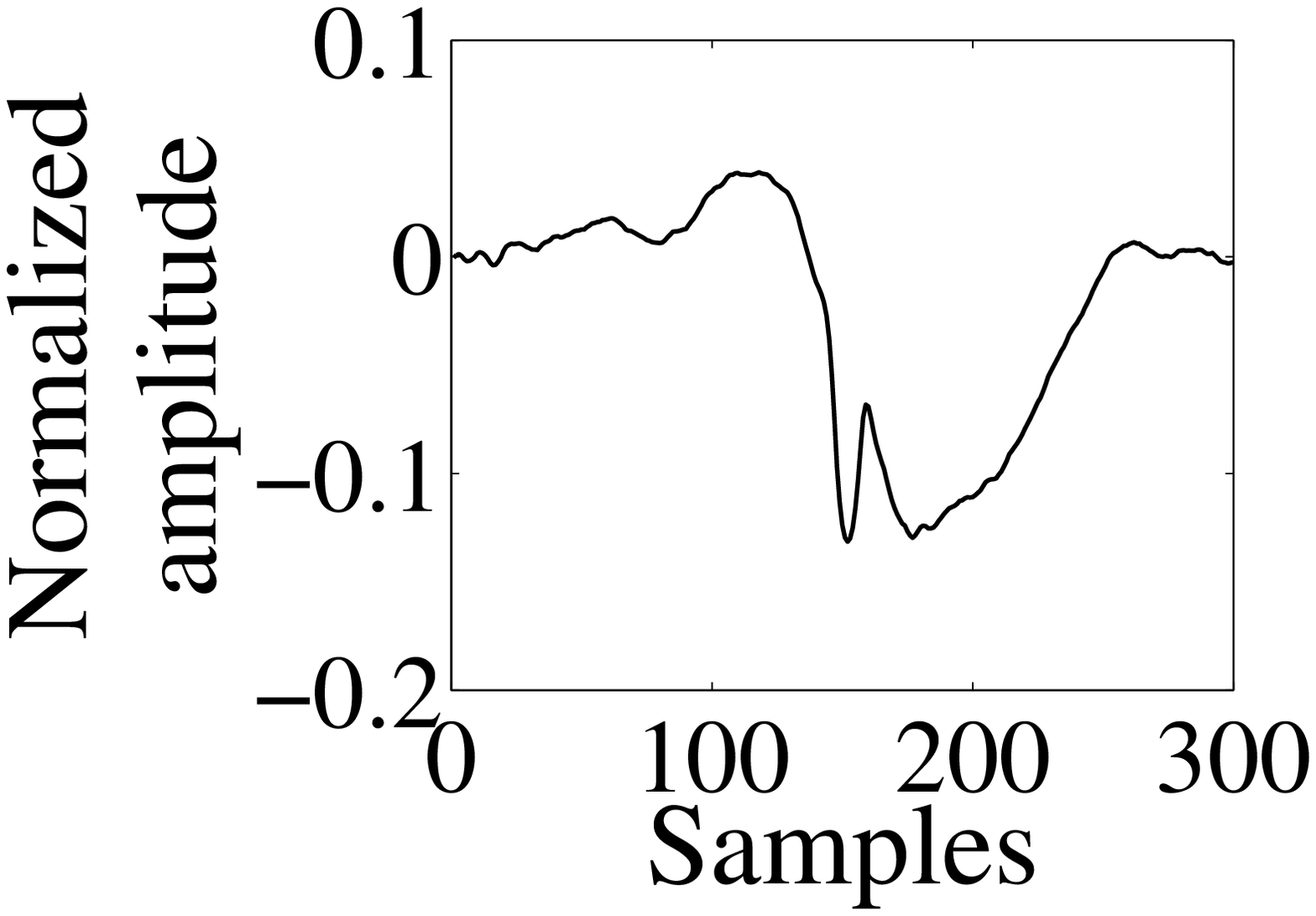}  \\
\includegraphics[width=3.5cm]{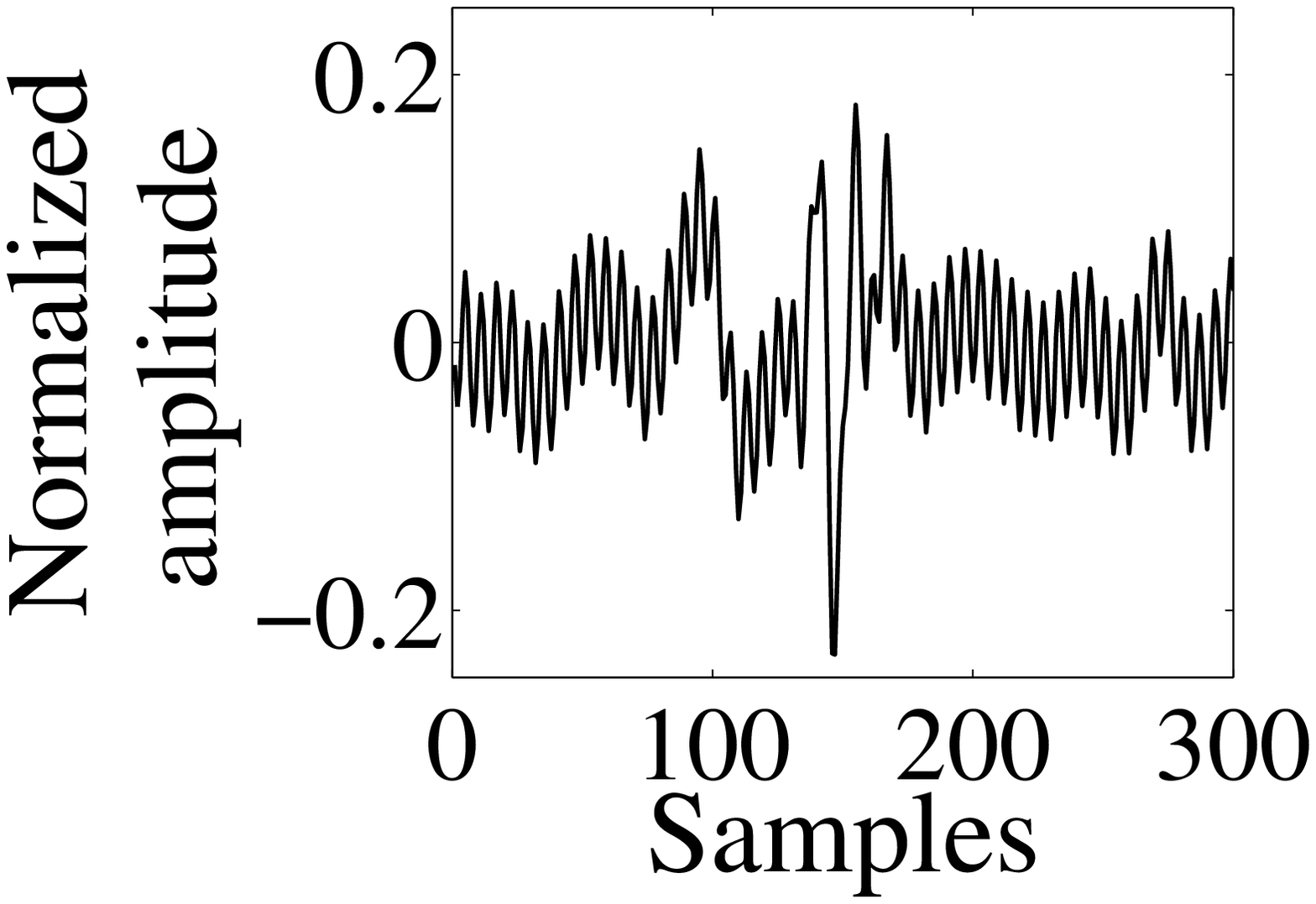}  & \includegraphics[width=3.5cm]{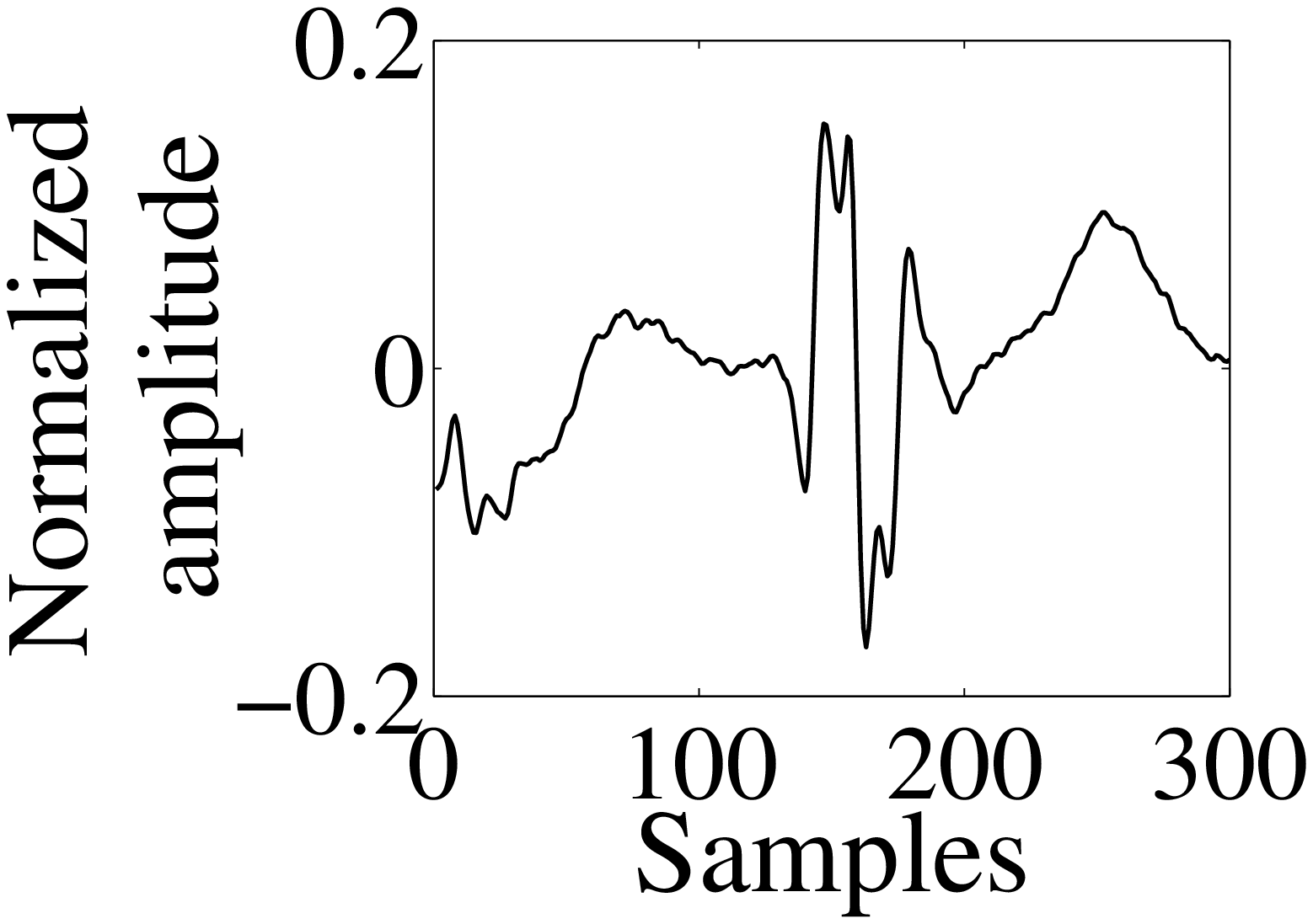}\\
(c) & (d) \\
\end{tabular}
\caption{(a) Typical normal beat; (b) Typical PVC beat; Three most frequently used (top to bottom in decreasing order) atoms in respective dictionaries corresponding to (c) normal beats and (d) PVC beats.}
\label{atoms}
\end{centering}
\end{figure}

\subsection{Dictionary Size}

In the proposed dictionary-based classification/compression approach, we trained an overcomplete dictionary for each of normal and PVC classes using K-SVD algorithm. In this regard, the dictionary size assumed importance, as (i) smaller size required less computation, and (ii) larger column size led to sparser representation, both of which properties are desirable. However, sparsity saturates with increasing column size, and has negligible effect on classification performance beyond certain threshold \cite{HC14}. Accordingly, we seek to choose the smallest dictionary that provides acceptable level of sparsity. Empirically, ``good" overcomplete dictionaries have been shown to possess a ratio of column to row size between approximately 2 and 5 \cite{ksvd}. Fortunately, we achieved satisfactory classification and compression performance, even while operating at the lower limit of the said ratio range. In particular, recalling that the row size equals the signal vector length of 301, we made use of dictionaries of size 301$\times$600.

\begin{figure}[t!]
%\centering
%\hspace*{-20pt}
\begin{tabular}{c}
\includegraphics[width=0.5\textwidth]{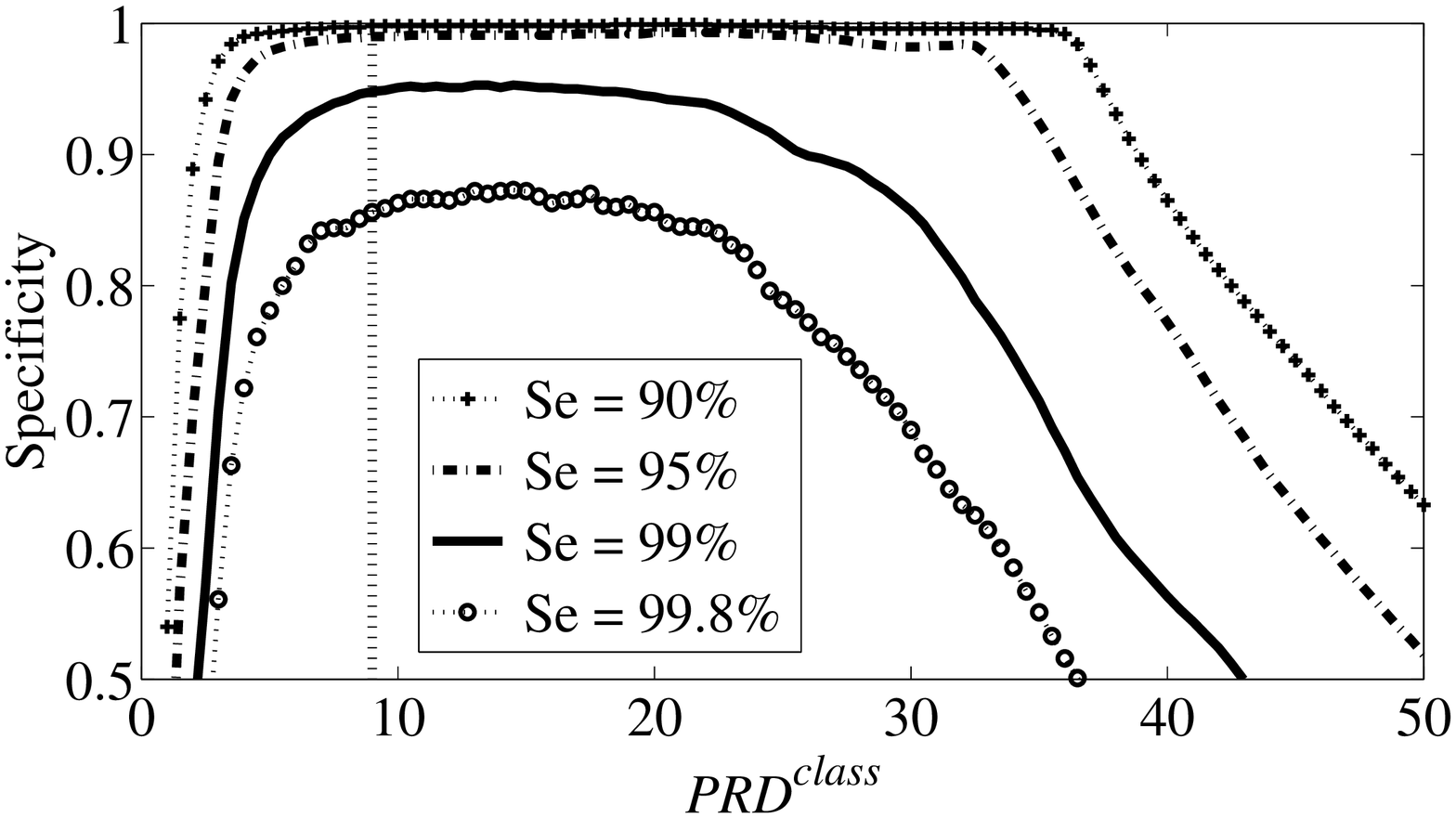} \\
(a)\\
\includegraphics[width=0.5\textwidth]{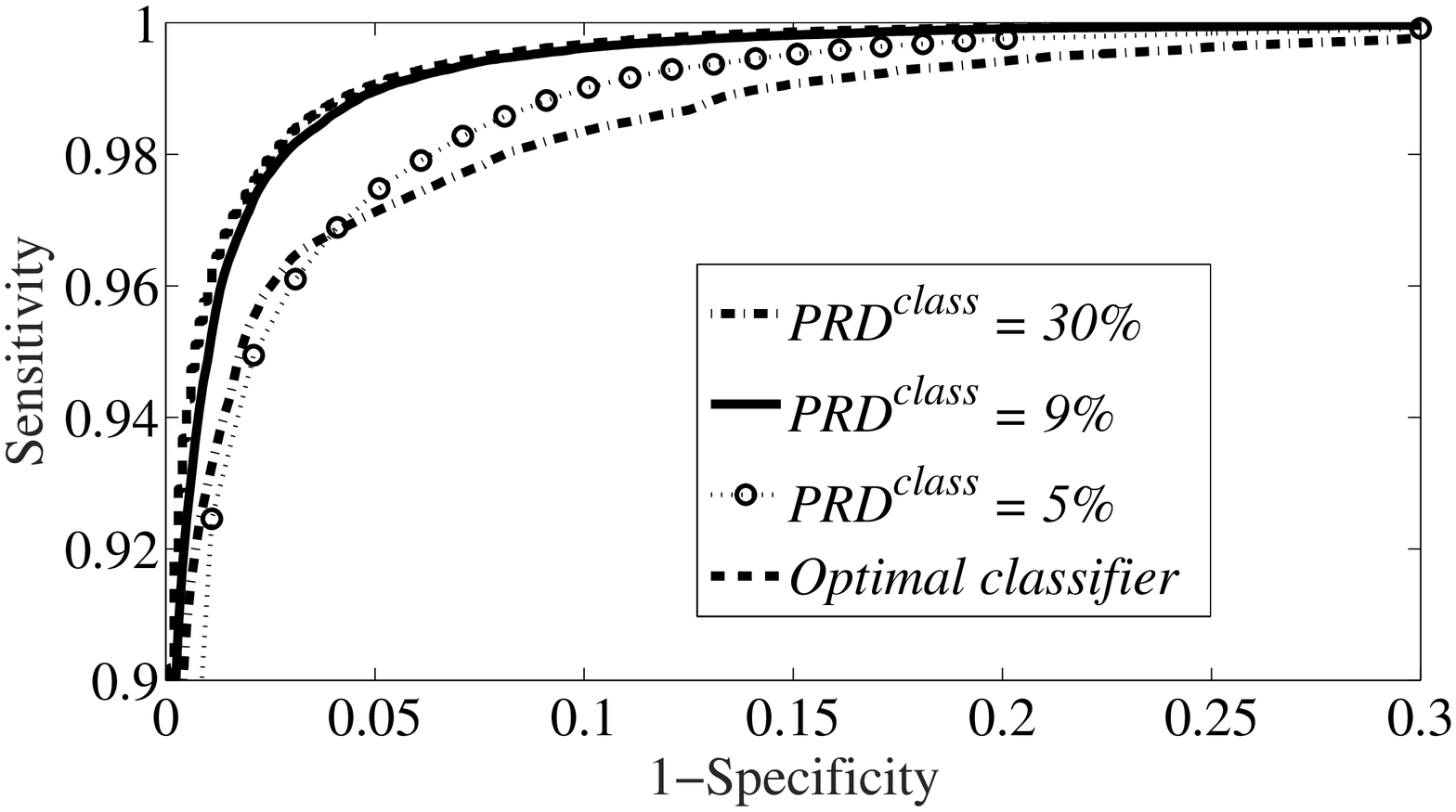} \\
(b)\\
\includegraphics[width=0.5\textwidth]{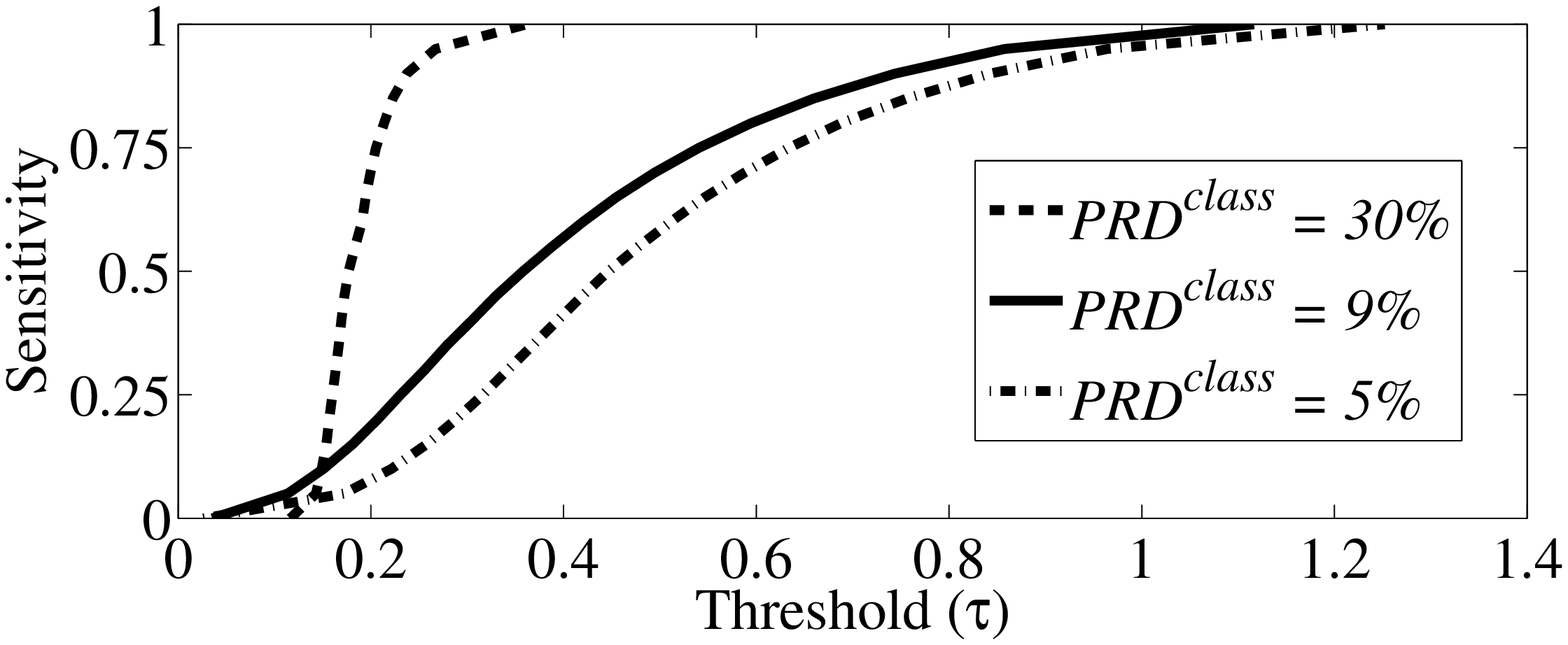} \\
(c)
\end{tabular}
\caption{(a) Specificity versus $PRD^{class}$ for various sensitivity values; (b) ROC ($Se$ versus $1-Sp$) curves at various $PRD^{class}$ values as well as for the optimal classifier; (c) Sensitivity versus threshold ($\tau$) at various $PRD^{class}$ values.}
\label{classifier_perf}
\end{figure}
%
%\begin{figure*}[]
%\begin{centering}
%\begin{tabular}{cc}
%\includegraphics[width=0.5\textwidth]{Classify_RsqComp} 
% & \includegraphics[width=0.5\textwidth]{Sp_plot_new} \\
%  
%(a) & (b) \\
%\end{tabular}•
%\caption{Plot of original and recovered signal and their corresponding error for PRD of 9\%, 5\% and 2\% respectively}
%\label{rec}
%\end{centering}
%\end{figure*}

%%%%%%%%%%%%%%%%%%%%%%%%%%%%%%%%%%%%%%%%%%%%%%%%%%%%%%%
\section{Experimental Results}
\label{sec:results}

In this section, we present experimental results, and performance analysis for the classification subsystem and the compression subsystem separately, as well as for the overall system. To this end, we made use of MIT/BIH Arrhythmia database  \cite{physionet}, adopted patient specific partitioning, evaluated the performance of our dictionary based method according to Proposal-1, Proposal-2 and Proposal-3, and compared with the performance of known algorithms, when relevant. For normal and PVC classes, separate dictionaries were obtained based on the training set, while the performance was evaluated using the test set. See Figure \ref{atoms} for a typical normal beat, a typical PVC beat as well as the three most frequently used atoms of each dictionary. Notice the differing beat morphologies, and how those are captured by the depicted dictionary atoms. For simulations, we used MATLAB v.2014b on a desktop computer with an Intel core i7 3.4 GHz 64-bit processor with 16 GB of memory, and required approximately 0.6 milliseconds to complete both classification and compression, which is several orders faster than real time, and indicates suitability of our system for practical deployment. Now we turn to reporting our results, beginning with classification performance.

\subsection{Classification Performance}

As mentioned earlier, classification performance depends on the reconstruction fidelity target $PRD^{class}$, internal to classification subsystem, which we choose first.
 
\subsubsection{Key tradeoffs and choice of $PRD^{class}$}

To this end, we studied the relationship among sensitivity, specificity and $PRD^{class}$ of the classifier for Proposal-3. Specifically, we plotted in Figure \ref{classifier_perf}a the tradeoff between specificity and $PRD^{class}$ at various sensitivity levels. At the sensitivity (reliability) target of 99\%, we observed specificity to be maximized at around $PRD^{class} = 9\%$. To appreciate the phenomenon from a different perspective, we 
plotted in Figure \ref{classifier_perf}b various ROC ($Se$ versus $1-Sp$) curves. There we found that the optimal ROC curve, obtained by varying $PRD^{class}$, is well approximated by the ROC curve at the fixed value $PRD^{class} = 9\%$. Accordingly, we  aimed at achieving a target $PRD^{class}$ of $9\%$.

As anticipated in Sec. \ref{sec:soln_class}, we observed in Figure \ref{classifier_perf}a that specificity indeed exhibits steep rise, near constancy (plateau) and steep fall as $PRD^{class}$ increases while keeping sensitivity levels fixed. Equivalently, in Figure \ref{classifier_perf}b, significantly lower (5\%) and higher (30\%) values of $PRD^{class}$ compared to the target $9\%$ lead to poor approximation of the optimal ROC. As an interesting aside, we noticed in Figure \ref{classifier_perf}a that the plateau region shrinks with increasing sensitivity. Further, recall that various levels of sensitivity $Se \in [0,1]$, and hence various points on the ROC, are obtained by varying a threshold $\tau$ on the sparsity ratio (see Sec. \ref{sec:soln_class}). Plotting $\tau$ versus $Se$ in Figure \ref{classifier_perf}c, we noticed that the range of $\tau$ shrinks, as the $PRD^{class}$ increases.

\begin{table*}[t!]
\renewcommand{\arraystretch}{1.3}
\centering
\begin{threeparttable}

\resizebox{160mm}{!}{
\begin{tabular}{C{4.5cm}ccC{6cm}C{6cm}}
\hline

\multirow{2}{*}{Methods} & \multirow{2}{*}{\begin{tabular}[c]{@{}l@{}}Se\\ (\%)\end{tabular}} & \multirow{2}{*}{\begin{tabular}[c]{@{}l@{}}Sp\\ (\%)\end{tabular}} & \multicolumn{2}{c}{Training Data} \\ \cline{4-5}
                        &                                                                    &                                                                    & Basic      & Patien-Specific      \\ \hline \hline

%Methods                      &  \begin{tabular}[c]{@{}c@{}}Se\\ (\%)\end{tabular} & \begin{tabular}[c]{@{}c@{}}Sp\\ (\%)\end{tabular}    & Train/test data + patient-specific training from test set  \\ \hline \hline

Y. H. Hu {\em et al.}, 1997 \cite{palreddy}  &  82.6                      & 97.1     &  Training records in Partition-1\tnote{*} & First 2.5 min of each test record in Partition-1 \\

P. de Chazal {\em et al.}, 2006 \cite{chazal} &  94.3          & 97            & Training records in Partition-3\tnote{*} & First 500 beats of each test record in Partition-3                                                                             \\ 

W. Jiang {\em et al.}, 2007 \cite{jiang}   &  86.6                                             & 99.3                                                                                        & Training records in Partition-2\tnote{*} & First 5 min of each test record in Partition-2  \\

T. Ince {\em et al.}, 2009 \cite{ince}       &  84.6             & 98.7           &  245 representative beats from training data of Partition-2\tnote{*} & First 5 min of each test record in Partition-2                             \\

M. Llamedo {\em et al\tnote{1}.}, 2012 \cite{Llamedo}  & 93    & 100   & Training records in Partition-3\tnote{*} & 12 annotated beats of each test record in Partition-3                                                                                        \\

%\begin{tabular}[c]{@{}c@{}}  Z. Zhang {\em et al.}, \\ 2014 \cite{zzhang}.  \end{tabular}       & Set-2\tnote{*} + No patient-specific training & \begin{tabular}[c]{@{}c@{}} 45843 (50.9\%) \\ 44283 (49.1\%) \end{tabular} & \begin{tabular}[c]{@{}c@{}}3788 (54\%) \\ 3220 (46\%)\end{tabular}   &  \begin{tabular}[c]{@{}c@{}}Ad hoc \\ random split\tnote{2} \end{tabular}  & 85.5                                            & 88.9                              \\ 

S. Kiranyaz {\em et al.}, 2016 \cite{kiranyaz}  & 93.9   & 98.9       & 245 representative beats from training data of Partition-2\tnote{*} & First 5 min of each test record in Partition-2                                             \\ \hline

%\multirow{3}{*}{\begin{tabular}[c]{@{}c@{}} \\ \\ {\bf Proposed} \\{\bf dictionary-based}\\ {\bf algorithm} \end{tabular}} 
{\bf Our proposals} & & & \\
Proposal-1 &  99                                               & 97.8     &   Training records in Partition-4\tnote{*}  &  First 5 min of each test record in Partition-4  \\ 
Proposal-2              & 99                                               & 89.9    $\pm$ 3.8           &   22 of 44  subjects are randomly chosen to form test set\tnote{2} , rest for training. Experiment is repeated for 100 random selections & First 5 min of each test record  \\ 
      {\bf  Proposal-3 (Recommended)}        &  {\bf 99}                                      & {\bf 95.3 $\pm$ 3.3}                                                                                         &    {\bf  4 of 44 subjects are randomly chosen to form test set\tnote{3}, rest for training. Experiment is repeated for 100 random selections.} & {\bf First 5 min of each test record} \\ \hline
\end{tabular}}

\begin{tablenotes}
\item[*] Details of Partition-1, Partition-2, Partition-3 and Partition-4 are given in Table \ref{data_partition}.
\item[1] Performance figures for only MIT/BIH database are given here. Such figures for other databases are also available. 
%\item[2] Evaluated on 10 random divisions and geometric mean performance is reported.
\item[2] Each test set has 45\%--55\% of total PVC beats; no constraint on normal beats.
\item[3] {\bf Each test set has 10\%--20\% of total PVC beats; no constraint on normal beats.}
\end{tablenotes}
 
\caption{Performance comparison of proposed algorithm with competing algorithms on MIT/BIH arrhythmia database.}
\label{comparison}
   \end{threeparttable}
\end{table*}

\subsubsection{Performance statistics}

Now, operating at $PRD^{class}$ = 9\%, we compared the performance of the proposed classifier with various algorithms that adopted patient specific evaluation scheme. Specifically, we report in Table \ref{comparison} sensitivity and specificity of existing classifiers and the proposed classifier along with specific information on training data. Recall that we set a sensitivity target of 99\% for our proposals. Now, comparing the peak performance, our Proposal-1 performs better than most of the reported algorithms. However, Proposal-1 represents an overly optimistic performance specific to Partition-4, and may not capture the performance variation due to randomly chosen partitioning. As a remedy, we reported the performance of our Proposal-2 and Proposal-3, where uncertainty is handled more realistically. Specifically, operating at the target sensitivity of 99\%, we reported the mean and standard deviation of specificity over 100 randomly chosen training and test sets. Subject to evenly split training and testing sets, the mean specificity obtained in Proposal-2 improves upon the peak specificity achieved in Proposal-1 in terms of fairness, although the former is significantly lower as expected. However, the notion of even split diverges from the practical situation, where significantly more data are available for training than testing. Accordingly, we recommend Proposal-3 that incorporates a realistic division with larger proportion of training data, as well as randomization. In this case, desirably, the mean specificity is higher, and the standard deviation is lower.

\begin{figure}[t!]
\centering
\includegraphics[width=0.5\textwidth]{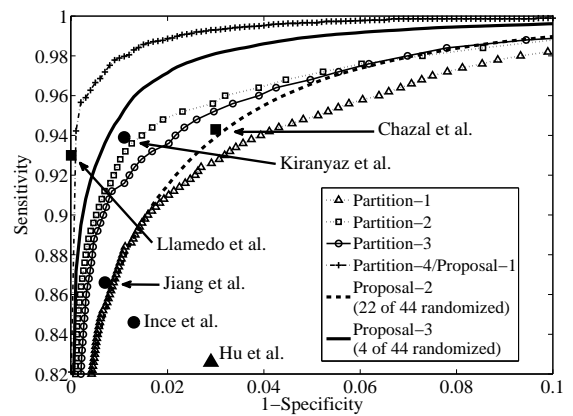} 
\caption{ROC of the classifier for various data partitions.}
\label{roc_all}
\end{figure}

%
%\begin{figure*}[t]
%\centering
%\begin{tabular}{ccc}
%\includegraphics[width=3.8cm]{CompRandomized} &
%\includegraphics[width=5cm]{ROC_set1} &
%\includegraphics[width=5cm]{ROC_set2} &
%\includegraphics[width=5cm]{ROC_set3} \\
%(a) & (b) & (c) \\
%\end{tabular}•
%\caption{ROC of the classifier for various sets of training data }
%\label{roc}
%\end{figure*}

So far, we furnished in Table \ref{comparison} pairs of sensitivity and specificity at the operating point of various algorithms. However, this information does not allow us to compare between those algorithms. Consequently, we used ROC curves to indicate the performance of our classifier across admissible sensitivity and specificity values (Figure \ref{roc_all}). In particular, we plotted ROC curves for Proposal-1, Proposal-2 and Proposal-3 (recommended). Alongside, ROC curves of our classifiers evaluated on Partition-1, Partition-2, Partition-3 and Partition-4 (Proposal-1) are also plotted. As ROCs of existing algorithms remain unavailable, we could only locate their operating points on the same plot. Encouragingly, Proposal-3 offers significant improvement over Proposal-2 as well as the peak performances of majority of reported results evaluated on hand-picked sets.

\subsection{Compression Performance}

%\input{Fig_Compression}
%We achieved beat compression by encoding only the quantized amplitude and the location of each of the non-zero entries of the sparse dictionary coefficients subject to reconstruction fidelity constraint $PRD^{compr}$. Recall from Sec \ref{sec:soln_compr}, the number of non-zero elements of dictionary coefficients and hence the compression performance is dictated by the intermediate reconstruction fidelity $PRD^{int}$. We now choose $PRD^{int}$ that maximizes compression ratio while maintaining desired $PRD^{compr}$.

Recall from Sec \ref{sec:soln_compr} that compression performance is determined by the number of non-zero elements in the dictionary coefficients, which is in turn dictated by the intermediate reconstruction fidelity $PRD^{int}$. We now choose $PRD^{int}$ that maximizes compression ratio while maintaining desired $PRD^{compr}$.

%\begin{figure}[t]
%\centering
%\includegraphics[width=3in]{CompRatioNewEdited}\\
%(a)  \\
%\includegraphics[width=3in]{BandwidthUsageNewEdited}\\
%(b) \\
%\caption{(a) Plot of compression ratio versus $R^2$ score for dictionary sizes 300 and 600; (b) Plot of bandwidth usage versus $R^2$ score.}
%\label{cr}
%\end{figure}

%\begin{figure}[]
%\centering
%\includegraphics[width=3in]{TempNorPVC}
%\caption{Compression ratio versus reconstruction fidelity for Normal and PVC dictionaries.}
%\label{cr}
%\end{figure}
%
%
%\begin{figure}[]
%\centering
%\includegraphics[width=3in]{117Temp}
%\caption{Compression ratio versus reconstruction fidelity for Normal and PVC dictionaries.}
%\label{cr}
%\end{figure}
%
%
%
%\begin{figure}[]
%\centering
%\includegraphics[width=3in]{117NoTS}
%\caption{Compression ratio versus reconstruction fidelity for Normal and PVC dictionaries.}
%\label{cr}
%\end{figure}

\begin{figure}[t]
\centering
\includegraphics[width=3in]{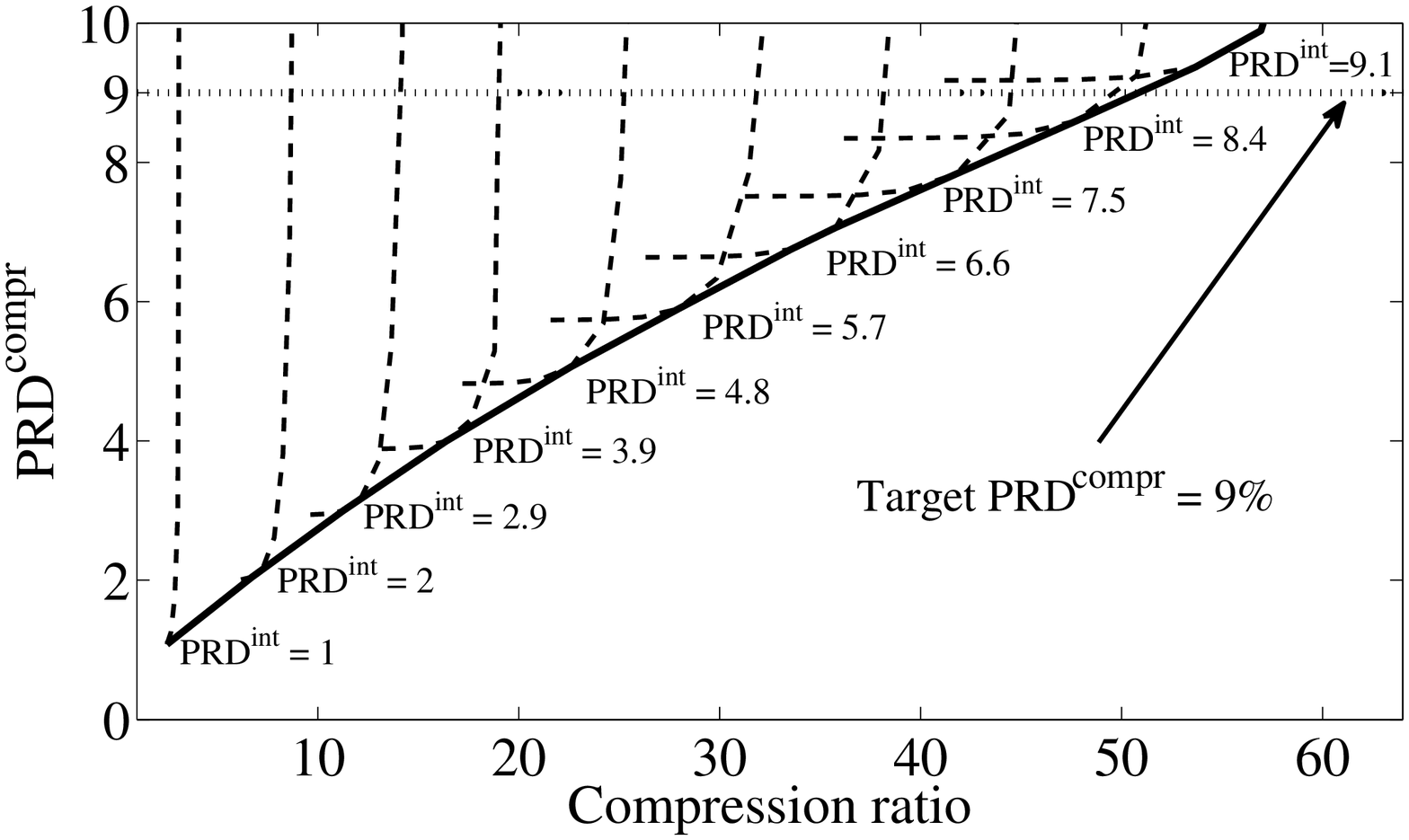}
(a)
\includegraphics[width=3in]{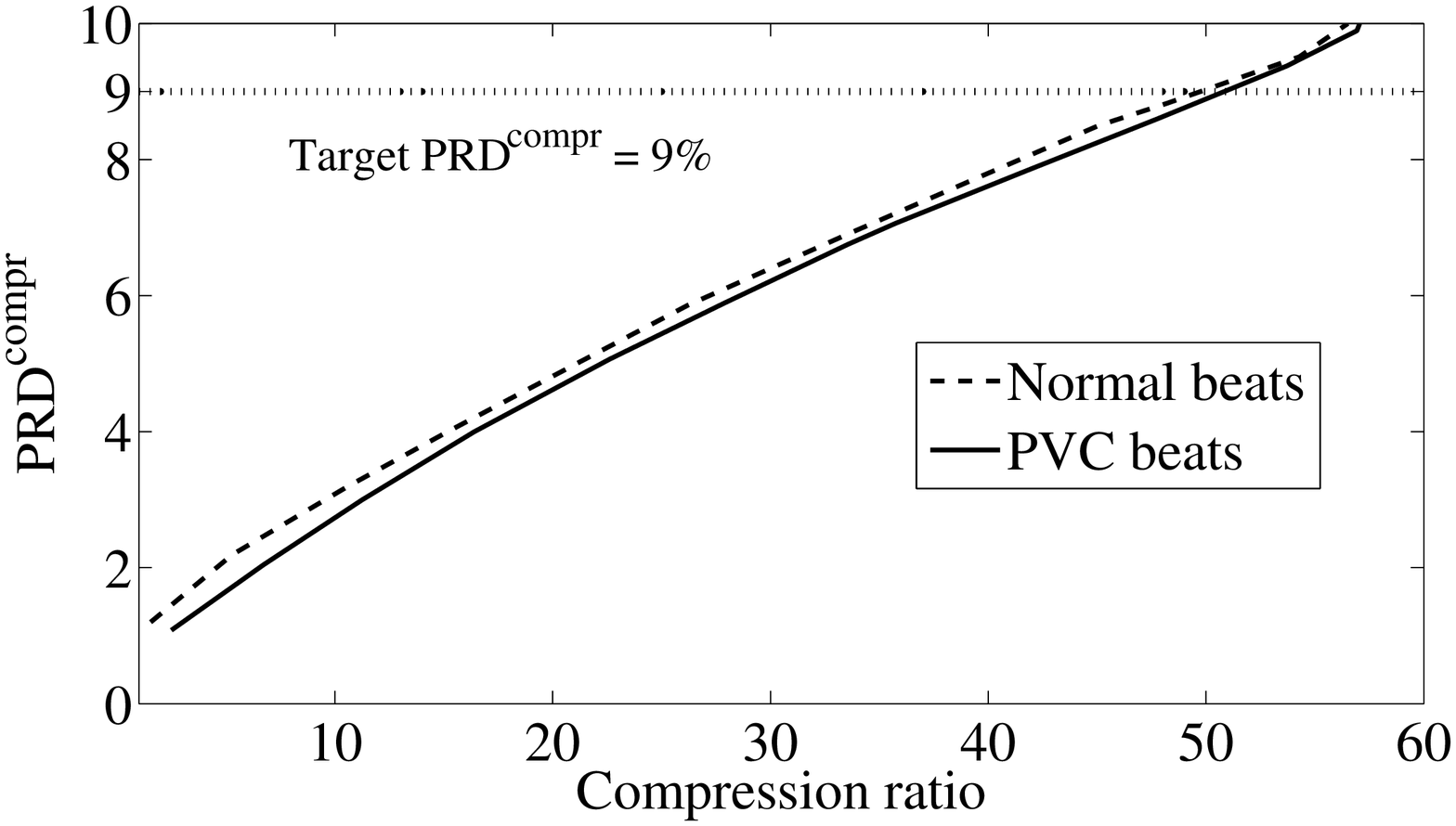}
(b)
\caption{Tradeoff between compression ratio and $PRD^{compr}$ (a) for various $PRD^{int}$ values corresponding to PVC beats; (b) envelope of the compression ratio versus $PRD^{compr}$ curve for normal and PVC beats.}
\label{cr}
\end{figure}

\subsubsection{Key tradeoffs and choice of $PRD^{int}$}

To this end, we first considered Proposal-3, and plotted $PRD^{compr}$ versus compression ratio for various $PRD^{int}$ values for  PVC beats (Figure \ref{cr}a). As mentioned earlier, for a small $PRD^{int}$, the number of non-zero elements of dictionary coefficients is large. In this setting, for a small increase in quantization step size $\Delta$, which produces a small increment in compression ratio $\beta_V$, quantization error from all those coefficients accumulate to result in a steep increment in $PRD^{compr}$. In contrast, for a large $PRD^{int}$, non-zero coefficients are few, and hence a similar increase in $\Delta$, producing a similarly small increment in $\beta_V$, now allows accumulation of quantization error from relatively few coefficients, resulting in only a gradual increment in $PRD^{compr}$. So, to plot the optimal curve of $PRD^{compr}$ versus compression ratio, we took the envelop of $PRD^{compr}$ versus compression ratio curves for various $PRD^{int}$ values. At this point, recall from Table \ref{PRD} that to preserve the diagnostic integrity of the ECG signal we should operate at at least $PRD^{compr}= 9\%$, which is indicated by the dotted line. For the choice $PRD^{compr} = 9\%$, we observed that $PRD^{int} \approx 8.8\%$ maximizes the compression ratio. The above steps were then repeated for normal beats, and optimal $PRD^{int} \approx 8.8\%$ was again observed. At this point, optimal $PRD^{compr}$ versus compression ratio curves for both PVC and normal beats were presented on the same plots in Figure \ref{cr}b. Those curves are similar with PVC beats allowing slightly higher compression ratio for any $PRD^{compr}$. 

%As anticipated higher $PRD^{compr}$ scores lead to significant compression in either dictionary due to the sparseness of the representation
%Note that the above steps were performed for both PVC and normal beats, although plots for only PVC beats are presented in Figure \ref{cr}a as representatives.
%Note that the above steps were performed for both PVC and normal beats, although plots for only PVC beats are presented in Figure \ref{cr}a as representatives.

%corresponding to PVC beats.

%Yet, to preserve the diagnostic integrity of the ECG signal, recall from Table \ref{PRD} that we should operate at a $PRD^{compr}$ no greater than 9\% as indicated by the dotted line the. In this context, we clamp the minimum compression ratio to one, because one may always send the entire signal to avoid $\beta<1$. 
%\begin{table}[]
%\centering
%\caption{My caption}
%\label{my-label}
%\begin{tabular}{ll}
%Method   & Compression Ratio \\
%22 of 44 &                   \\
%4 of 44  &                  
%\end{tabular}
%\end{table}

\begin{table}{}
\renewcommand{\arraystretch}{1.2}
\centering
\begin{threeparttable}

    	\begin{tabular}{ccc}
    	\toprule
	\multirow{2}{*}{Proposal} & \multicolumn{2}{c}{Compression ratio} \\
			   & $\beta_N$  & $\beta_V$   \\ \midrule
	Proposal-1\tnote{*}  & 51.8 &  53.1 \\
	Proposal-2\tnote{**} &        45.6 $\pm$ 2.8   & 39.1 $\pm$ 3.6  \\
	\bf {Proposal-3}\tnote{**}   &	    \bf{49.7 $\pm$ 6.5}  &  \bf{50.8 $\pm$ 8.3}  \\ \bottomrule
	\end{tabular}

\begin{tablenotes}
\item[*] Deterministic evaluation using Partition-4.
\item[**]  MCCV over 100 iterations. 
\end{tablenotes}

\caption{Compression ratio comparison.}
\label{table:CR}
\end{threeparttable}

\end{table}

\subsubsection{Performance statistics}

Compression performance too depends on the partitioning between training and test data.
To remove such dependency, we again adopted MCCV approach to evaluate our compression algorithm and reported the performance statistics for Proposal-1, Proposal-2 and Proposal-3 (Table \ref{table:CR}). Specifically, operating at $PRD^{compr}= 9\%$, we reported the mean and standard deviation of $\beta_N$ and $\beta_V$, compression ratios corresponding to normal beats and PVC beats, respectively. Not unexpectedly, with larger training data (Proposal-3), mean compression performance increased. Interestingly, standard deviation of compression performance also increased. 

Here, unlike in the case of classification, the proposed compression technique cannot fairly be compared with state of the art algorithms. To appreciate this, note that certain algorithms achieve high compression ratio for a signal consisting of several beats by stacking such beats before compressing \cite{tai}. In contrast, we avoid stacking to prevent delays. Further, specific fixed partitioning is sometimes chosen so as to maximize reported compression ratio \cite{fira1, adamo}. As mentioned earlier, such unrandomized results cannot be used as guides for practical system design.

\subsection{System Performance}

We now present the overall system performance in terms of savings in bandwidth cost. Operating at the target reliability of $Se = 99\%$ and reconstruction fidelity of $PRD^{compr} = 9\%$, the proposed beat-trio communication system would achieve about  57.6\% savings of original bandwidth if classification alone was used. Using compression alone, bandwidth savings is increased to 97.99\%, while ignoring the communication overheads. Using both classification and compression, the proposed method achieved 99.15\% saving in bandwidth usage, which translates to a proportionate savings in the operating cost.

\begin{figure}[]
\centering
\includegraphics[width=0.5\textwidth]{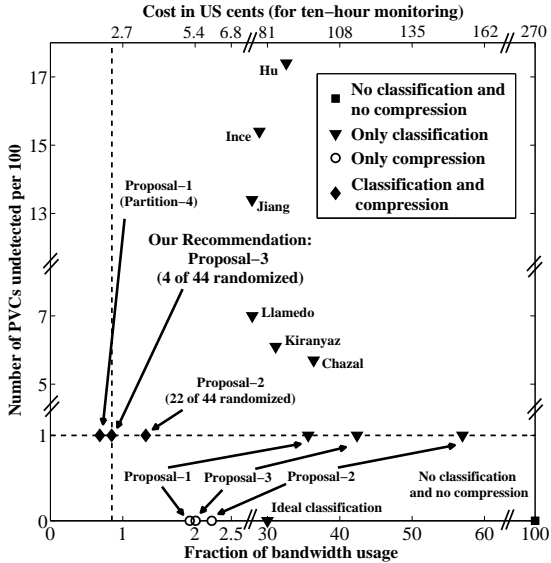}
\caption{Comparison of various classifiers in the context of telecardiology.}
\label{comp}
\end{figure}

%\subsubsection{Cost comparison in telecardiological context}

Now, let us revisit the representative scenario presented in Section \ref{sec:motivation}, and recall that conventional telecardiology costs about US\$ 2.7 for 10-hour ECG monitoring. Against this reference cost, in Figure \ref{comp} we graphically depicted the performance of our system as well as other reported algorithms in the telecardiological context. Specifically, we located various systems in a reliability versus bandwidth/cost plane. In $y$-axis, we plot the complement of reliability ($1-Se$), i.e., the number of PVCs undetected per one hundred beats, and in $x$-axis, the bandwidth usage (bottom) as well as the cost (top) for ten-hour monitoring. Now, adopting beat-trio transmission, and assuming a PVC prevalence rate $\rho$ = 10\%, we plotted reliability versus the bandwidth cost for various rival algorithms, and our Proposal-1, Proposal-2 and Proposal-3, without as well as with compression. Employing only classification, our recommended proposal required only 42.4\% of bandwidth. Notice that a number of reported classifiers did not perform close to the reliability target of $Se=99\%$, i.e., one undetected PVC in one hundred as indicated by horizontal dashed line. The nearest in this respect, the classifier proposed by Chazal {\em et al.} \cite{chazal}, requires 36.4\% of the reference bandwidth, while missing about six PVCs in one hundred beats, i.e., operating at a rate six-fold higher than the target. Using only compression, our proposal reduced the bandwidth requirement to only 2\% of the original bandwidth. However, such a scheme would burden the medical professional with processing entire record for diagnosis. In comparison, the proposed classifier employing both classification and compression would not only reduce the bandwidth requirement but also assist medical professionals by localizing potential anomalies. Specifically, our system would use only 0.85\% of the original bandwidth, achieving additional 98\% and 57.5\% savings over the bandwidth required for classification alone and compression alone, respectively. This would bring down the operating cost to US\textcentoldstyle~2.3. At this rate, the healthcare expenses of the household, mentioned in Section \ref{sec:motivation}, would be reduced to an affordable 0.46\% of the household income from the original 54\%. We believe that a drastic cost reduction of this scale should enable the targeted BOP communities to opt for continuous monitoring service without severe economic burden.

\section{Discussion}
\label{sec:disc}

We conclude by summarizing our contributions, remarking on the anticipated user experience, and reflecting on broader impact of our work.

%
%We now discuss about various aspects pertinent to the proposed telecardiology system.

\subsection{Summary}

In this paper, we presented an ultra-low-cost POC service for PVC monitoring that ensures high accuracy. In particular, we proposed a dictionary-based technique that achieves high-sensitivity classification and high-fidelity compression. We demonstrated the efficacy of our method using Monte Carlo cross validation on the MIT/BIH arrhythmia database \cite{MCCV, physionet}. In particular, the three-way tradeoff between bandwidth, reliability and reconstruction fidelity was characterized. With a reliability target of at most one undetected PVC in one hundred beats, and a reconstruction fidelity of 9\% level of $PRD$, we achieved about forty-fold savings in bandwidth and the associated cost. Our service would cost only US\textcentoldstyle~2.3 for ten-hour monitoring, which, we believe, should be attractive to the economically marginalized. 

%Revisiting the representative scenario presented in Section \ref{sec:motivation}, employing the proposed POC service would bring down the cost from US\$2.7 to US\textcentoldstyle 6.8. Amounting to only about 1.36\% of the household income. Such an ultra-low-cost solution would enable the targeted BOP communities to avail the continuous monitoring service.

\subsection{User Experience}

While using our service, the experience of users (both subjects and medical professionals) is anticipated to remain essentially the same as that associated with conventional telecardiology. Specifically, at the subject end, the same transducers are still used to collect the ECG signals from the patient. From the medical professionals' perspective, the inference has to be made from the electronic records at essentially the same quality ($PRD \le 9\%$, from Table \ref{PRD} \cite{wdd}) as the gold (quality) standard of unprocessed signals. In fact, the time and effort required of the medical professional are anticipated to be less than that in the traditional situation, as the proposed method automatically identifies PVCs and presents only delimited anomalous beats. In a nutshell, subjects familiar with convectional telecardiology would require no additional training, whereas  medical professionals would only need to focus on the presented beats (beat-trios), and ignore blank spaces, which would just indicate normal (uninformative) beats.

%Thus, the medical professionals are facilitated to focus on clinically relevant beats, rather than visually process entire record to identify PVCs.

%While assuring the quality of service of the proposed system, bedside monitoring should be treated as the gold standard \cite{kastania}. We strive to achieve clinical decisions that are statistically indistinguishable from the outcome based on bedside system. In the proposed POC service, signals which are detected as normal, and not adjacent to a PVC beat are no longer conveyed to diagnostic center. Hence, there remains no opportunity for the medical professionals to correct a mistake if the underlying beat is anomalous. To this end, we set classifier sensitivity requirement greater than 99\% to limit such adverse effects. Further, we ensure PRD to be less than 9\%, that indicate good signal reconstruction. Thus, with a minor compensation on sensitivity and PRD values, proposed scheme achieves significant savings when compared to gold standard.

\subsection{Broader Impact}

Monitoring of PVCs is clinically significant in broader scenarios than considered so far. Specifically, high PVC burden could presage adverse heart conditions even in individuals without prior structural heart disease \cite{ng}. In such contexts, our technique with slight modifications could facilitate preventive care. Further, apart from PVCs, the proposed dictionary-based method could be extended to other anomalous indicators such as supraventricular arrhythmias and atrial fibrillation \cite{page}. In addition, incorporating medical professionals' feedback and adaptively learning personalized dictionaries could potentially improve both classification and compression performance levels \cite{mairal}.

\section*{Acknowledgment}

%Authors are thankful to Laxminarayana Anumandla (Cardiothoracic Surgeon, Maxcare Hospital, Warangal, India) for helpful discussions on clinical aspects. 
This work was partially supported by the Department of Electronics and Information Technology (DeitY), Govt. of India, under the Cyber Physical Systems Innovation Project: 13(6)/2010- CC\&BT.

%\input{referencesNew}
%\begin{multicols}{0}
\bibliographystyle{elsarticle-num}
%\bibliography{Ref_AIM_new}

\begin{thebibliography}{10}
\expandafter\ifx\csname url\endcsname\relax
  \def\url#1{\texttt{#1}}\fi
\expandafter\ifx\csname urlprefix\endcsname\relax\def\urlprefix{URL }\fi
\expandafter\ifx\csname href\endcsname\relax
  \def\href#1#2{#2} \def\path#1{#1}\fi

\bibitem{WHO}
World health organization, fact sheet on cvds (fact sheet n${}^{0}$317), 2013.

\bibitem{podrid}
P.~J. Podrid, P.~R. Kowey, Cardiac arrhythmia: mechanisms, diagnosis, and
  management, Lippincott Williams \& Wilkins, 2001.

\bibitem{gertsch}
M.~Gertsch, The ECG: a two-step approach to diagnosis, Springer Science \&
  Business Media, 2003.

\bibitem{hinkle}
L.~Hinkle, S.~Carver, D.~Argyros, The prognostic significance of ventricular
  premature contractions in healthy people and in people with coronary heart
  disease., Acta cardiologica (1974) 5.

\bibitem{ng}
G.~A. Ng, Treating patients with ventricular ectopic beats, Heart 92~(11)
  (2006) 1707--1712.

\bibitem{poverty} \url{http://hdr.undp.org/en/content/population-living-below-125-ppp-day}

\bibitem{west}
D.~West, How mobile devices are transforming healthcare, Issues in technology
  innovation 18~(1) (2012) 1--11.

\bibitem{BOP}
C.~K. Prahalad, The Fortune at the Bottom of the Pyramid, Pearson Education
  India, 2006.

\bibitem{frugal}
N.~Kumar, P.~Puranam, Frugal engineering: An emerging innovation paradigm, Ivey
  Business Journal 76~(2) (2012) 14--16.

\bibitem{palreddy}
Y.~H. Hu, S.~Palreddy, W.~J. Tompkins, A patient-adaptable ecg beat classifier
  using a mixture of experts approach, IEEE transactions on biomedical
  engineering 44~(9) (1997) 891--900.

\bibitem{chazal}
P.~de~Chazal, R.~B. Reilly, A patient-adapting heartbeat classifier using ecg
  morphology and heartbeat interval features, IEEE Transactions on Biomedical
  Engineering 53~(12) (2006) 2535--2543.

\bibitem{Llamedo}
M.~Llamedo, J.~P. Mart{\'\i}nez, An automatic patient-adapted ecg heartbeat
  classifier allowing expert assistance, IEEE Transactions on Biomedical
  Engineering 59~(8) (2012) 2312--2320.

\bibitem{melgani}
F.~Melgani, Y.~Bazi, Classification of electrocardiogram signals with support
  vector machines and particle swarm optimization, IEEE Transactions on
  Information Technology in Biomedicine 12~(5) (2008) 667--677.

\bibitem{jiang}
W.~Jiang, S.~G. Kong, Block-based neural networks for personalized ecg signal
  classification, IEEE Transactions on Neural Networks 18~(6) (2007)
  1750--1761.

\bibitem{ince}
T.~Ince, S.~Kiranyaz, M.~Gabbouj, Automated patient-specific classification of
  premature ventricular contractions, in: 2008 30th Annual International
  Conference of the IEEE Engineering in Medicine and Biology Society, IEEE,
  2008, pp. 5474--5477.

\bibitem{zzhang}
Z.~Zhang, J.~Dong, X.~Luo, K.-S. Choi, X.~Wu, Heartbeat classification using
  disease-specific feature selection, Computers in biology and medicine 46
  (2014) 79--89.

\bibitem{kiranyaz}
S.~Kiranyaz, T.~Ince, M.~Gabbouj, Real-time patient-specific ecg classification
  by 1-d convolutional neural networks, IEEE Transactions on Biomedical
  Engineering 63~(3) (2016) 664--675.

\bibitem{bortolan}
G.~Bortolan, I.~Jekova, I.~Christov, Comparison of four methods for premature
  ventricular contraction and normal beat clustering, in: Computers in
  Cardiology, 2005, IEEE, 2005, pp. 921--924.

\bibitem{tai}
S.-C. Tai, C.~Sun, W.-C. Yan, A 2-d ecg compression method based on wavelet
  transform and modified spiht, IEEE Transactions on Biomedical Engineering
  52~(6) (2005) 999--1008.

\bibitem{zigel1}
Y.~Zigel, A.~Cohen, A.~Katz, Ecg signal compression using analysis by synthesis
  coding, IEEE Transactions on Biomedical Engineering 47~(10) (2000)
  1308--1316.

\bibitem{fira1}
C.~M. Fira, L.~Goras, An ecg signals compression method and its validation
  using nns, IEEE Transactions on Biomedical Engineering 55~(4) (2008)
  1319--1326.

\bibitem{adamo}
A.~Adamo, G.~Grossi, R.~Lanzarotti, J.~Lin, Ecg compression retaining the best
  natural basis k-coefficients via sparse decomposition, Biomedical Signal
  Processing and Control 15 (2015) 11--17.

\bibitem{jana}
S.~Jana, P.~Moulin, Optimality of klt for high-rate transform coding of
  gaussian vector-scale mixtures: Application to reconstruction, estimation,
  and classification, IEEE Transactions on Information Theory 52~(9) (2006)
  4049--4067.

\bibitem{vlodaver}
Z.~Vlodaver, R.~F. Wilson, D.~Garry, Coronary Heart Disease: Clinical,
  Pathological, Imaging, and Molecular Profiles, Springer Science \& Business
  Media, 2012.

\bibitem{cha}
Y.-M. Cha, G.~K. Lee, K.~W. Klarich, M.~Grogan, Premature ventricular
  contraction-induced cardiomyopathy a treatable condition, Circulation:
  Arrhythmia and Electrophysiology 5~(1) (2012) 229--236.

\bibitem{niwano}
S.~Niwano, Y.~Wakisaka, H.~Niwano, H.~Fukaya, S.~Kurokawa, M.~Kiryu,
  Y.~Hatakeyama, T.~Izumi, Prognostic significance of frequent premature
  ventricular contractions originating from the ventricular outflow tract in
  patients with normal left ventricular function, Heart 95~(15) (2009)
  1230--1237.

\bibitem{baman}
T.~S. Baman, D.~C. Lange, K.~J. Ilg, S.~K. Gupta, T.-Y. Liu, C.~Alguire,
  W.~Armstrong, E.~Good, A.~Chugh, K.~Jongnarangsin, et~al., Relationship
  between burden of premature ventricular complexes and left ventricular
  function, Heart Rhythm 7~(7) (2010) 865--869.

\bibitem{scheidt}
S.~Scheidt, J.~McGill, G.~Wilner, T.~Killip, Remote electrocardiography:
  clinical experience with telephone transmission of electrocardiograms, JAMA
  230~(9) (1974) 1293--1294.

\bibitem{varshney}
U.~Varshney, Pervasive healthcare and wireless health monitoring, Mobile
  Networks and Applications 12~(2-3) (2007) 113--127.

\bibitem{huang}
A.~Huang, C.~Chen, K.~Bian, X.~Duan, M.~Chen, H.~Gao, C.~Meng, Q.~Zheng,
  Y.~Zhang, B.~Jiao, et~al., We-care: an intelligent mobile telecardiology
  system to enable mhealth applications, IEEE journal of biomedical and health
  informatics 18~(2) (2014) 693--702.

\bibitem{leehj}
H.~J. Lee, S.~H. Lee, K.-S. Ha, H.~C. Jang, W.-Y. Chung, J.~Y. Kim, Y.-S.
  Chang, D.~H. Yoo, Ubiquitous healthcare service using zigbee and mobile phone
  for elderly patients, International journal of medical informatics 78~(3)
  (2009) 193--198.

\bibitem{Mukhopadhyay}
S.~K. Mukhopadhyay, S.~Mitra, M.~Mitra, A lossless ecg data compression
  technique using ascii character encoding, Computers \& Electrical Engineering
  37~(4) (2011) 486--497.

\bibitem{ICT}
W.~H. Lehr, Quality and Reliability of Telecommunications Infrastructure,
  Routledge, 2013.

\bibitem{steg}
P.~G. Steg, S.~K. James, D.~Atar, L.~P. Badano, C.~B. Lundqvist, M.~A. Borger,
  C.~Di~Mario, K.~Dickstein, G.~Ducrocq, F.~Fernandez-Aviles, et~al., Esc
  guidelines for the management of acute myocardial infarction in patients
  presenting with st-segment elevation, European heart journal (2012) ehs215.

\bibitem{purohit}
B.~C. Purohit, Private initiatives and policy options: recent health system
  experience in india, Health policy and planning 16~(1) (2001) 87--97.

\bibitem{jaroslawski}
S.~Jaros{\l}awski, G.~Saberwal, In ehealth in india today, the nature of work,
  the challenges and the finances: an interview-based study, BMC medical
  informatics and decision making 14~(1) (2014) 1.

\bibitem{physionet}
A.~L. Goldberger, L.~A. Amaral, L.~Glass, J.~M. Hausdorff, P.~C. Ivanov, R.~G.
  Mark, J.~E. Mietus, G.~B. Moody, C.-K. Peng, H.~E. Stanley, Physiobank,
  physiotoolkit, and physionet components of a new research resource for
  complex physiologic signals, Circulation 101~(23) (2000) e215--e220.

\bibitem{affordability}
L.~Ni{\"e}ns, E.~Van~de Poel, A.~Cameron, M.~Ewen, R.~Laing, W.~Brouwer,
  Practical measurement of affordability: an application to medicines, Bulletin
  of the World Health Organization 90~(3) (2012) 219--227.

\bibitem{household}
\href{http://www.nakono.com/tekcarta/databank/households-average-household-size}{[link]}.
\newline\urlprefix\url{http://www.nakono.com/tekcarta/databank/households-average-household-size}

\bibitem{NP}
J.~Neyman, E.~S. Pearson, On the problem of the most efficient tests of
  statistical hypotheses, in: Breakthroughs in Statistics, Springer, 1992, pp.
  73--108.

\bibitem{sornmo}
L.~S{\"o}rnmo, P.~Laguna, Bioelectrical signal processing in cardiac and
  neurological applications, Vol.~8, Academic Press, 2005.

\bibitem{wdd}
Y.~Zigel, A.~Cohen, A.~Katz, The weighted diagnostic distortion (wdd) measure
  for ecg signal compression, IEEE Transactions on Biomedical Engineering
  47~(11) (2000) 1422--1430.

\bibitem{elad}
M.~Elad, Sparse and Redundant Representations: From Theory to Applications in
  Signal and Image Processing, 1st Edition, Springer Publishing Company,
  Incorporated, 2010.

\bibitem{rubinstein}
R.~Rubinstein, M.~Zibulevsky, M.~Elad, Efficient implementation of the k-svd
  algorithm using batch orthogonal matching pursuit, CS Technion 40~(8) (2008)
  1--15.

\bibitem{ksvd}
M.~Aharon, M.~Elad, A.~Bruckstein, K-svd: An algorithm for designing
  overcomplete dictionaries for sparse representation, IEEE Transactions on
  signal processing 54~(11) (2006) 4311--4322.

\bibitem{haykin}
S.~Haykin, Communication systems, John Wiley \& Sons, 2008.

\bibitem{ye}
C.~Ye, B.~V. Kumar, M.~T. Coimbra, Heartbeat classification using morphological
  and dynamic features of ecg signals, IEEE Transactions on Biomedical
  Engineering 59~(10) (2012) 2930--2941.

\bibitem{krasteva}
V.~Krasteva, I.~Jekova, Qrs template matching for recognition of ventricular
  ectopic beats, Annals of Biomedical Engineering 35~(12) (2007) 2065--2076.

\bibitem{aami}
A.-A. EC57, Testing and reporting performance results of cardiac rhythm and st
  segment measurement algorithms, Association for the Advancement of Medical
  Instrumentation, Arlington, VA.

\bibitem{HC14}
B.~S. Chandra, C.~S. Sastry, S.~Jana, Reliable low-cost telecardiology:
  High-sensitivity detection of ventricular beats using dictionaries, in:
  e-Health Networking, Applications and Services (Healthcom), 2014 IEEE 16th
  International Conference on, IEEE, 2014, pp. 305--310.

\bibitem{MCCV}
S.~Arlot, A.~Celisse, et~al., A survey of cross-validation procedures for model
  selection, Statistics surveys 4 (2010) 40--79.

\bibitem{page}
R.~L. Page, W.~E. Wilkinson, W.~K. Clair, E.~A. McCarthy, E.~Pritchett,
  Asymptomatic arrhythmias in patients with symptomatic paroxysmal atrial
  fibrillation and paroxysmal supraventricular tachycardia., Circulation 89~(1)
  (1994) 224--227.

\bibitem{mairal}
J.~Mairal, F.~Bach, J.~Ponce, G.~Sapiro, Online dictionary learning for sparse
  coding, in: Proceedings of the 26th annual international conference on
  machine learning, ACM, 2009, pp. 689--696.

\end{thebibliography}

%\end{multicols}

\end{document}